\documentclass[11pt]{article} %

\usepackage{lmodern}
\usepackage{xspace}                                     %
\usepackage[protrusion=true,expansion=false]{microtype}  %
\usepackage{times}

\usepackage[style=alphabetic,natbib=true,maxnames=99,maxbibnames=99,maxcitenames=99,maxalphanames=99]{biblatex}
\bibliography{bibliography, references}
\usepackage{framed}
\usepackage{dsfont}
\usepackage{fullpage}
\usepackage{url}
\usepackage{amsmath,amsthm,amssymb,mathtools}
\usepackage[ruled]{algorithm2e} %
\usepackage[colorlinks,citecolor=blue,bookmarks=true,linktocpage]{hyperref}
\usepackage[capitalize]{cleveref}
\usepackage{aliascnt} 

\usepackage{MaryamsMacros}

\theoremstyle{definition}
\newtheorem{definition}{Definition}

\theoremstyle{plain}
\newtheorem{theorem}{Theorem}
\newaliascnt{thelemma}{theorem}
\newtheorem{lemma}[thelemma]{Lemma}
\crefname{lemma}{lemma}{lemmas}
\newaliascnt{theremark}{theorem}
\newtheorem{remark}[theremark]{Remark}
\crefname{remark}{remark}{remarks}

\usepackage{mleftright}
\newcommand{\eps}{\varepsilon}
\newcommand{\errprob}{\delta}
\newcommand\Lap[1]{\mathbf{Lap}\mleft(#1\mright)}
\newcommand\Poi[1]{\mathbf{Poi}\mleft(#1\mright)}
\newcommand\Bern[1]{\mathbf{Bern}\mleft(#1\mright)}

\newcommand\Ham[2]{\operatorname{Ham}\mleft(#1, #2\mright)}
\newcommand\totalvardist[2]{\operatorname{d}_{\rm{}TV}\mleft(#1, #2\mright)}
\newcommand{\accept}{\mathsf{accept}}
\newcommand{\reject}{\mathsf{reject}}
\newcommand{\invalid}{\bot}
\newcommand{\privacyeps}{\xi}
\newcommand{\testingeps}{\eps}
\newcommand{\accuracy}{\alpha}

\newcommand{\myparagraph}[1]{\smallskip\noindent\textbf{#1}}

\providecommand{\email}[1]{\href{mailto:#1}{\nolinkurl{#1}\xspace}}

\newif\ifcolt 

\title{Better Private Distribution Testing by Leveraging \\ Unverified Auxiliary Data}
\author{Maryam Aliakbarpour\thanks{Department of Computer Science \& Ken Kennedy Institute, Rice University. Email: \email{maryama@rice.edu}.}
\and Arnav Burudgunte\thanks{Purdue University. Email: \email{aburudgu@purdue.edu}.}
\and Cl\'ement L. Canonne\thanks{University of Sydney. Email: \email{clement.canonne@sydney.edu.au}. Supported by an ARC DECRA (DE230101329).}
\and Ronitt Rubinfeld\thanks{Computer Science and Artificial Intelligence Laboratory, MIT. Email: \email{ronitt@csail.mit.edu}. Supported by the NSF
TRIPODS program (award DMS-2022448) and CCF-2310818}}

\begin{document}

\coltfalse

\maketitle

\begin{abstract}%
  We extend the framework of augmented distribution testing (Aliakbarpour, Indyk, Rubinfeld, and Silwal, NeurIPS 2024) to the differentially private setting. This captures scenarios where a data analyst must perform hypothesis testing tasks on sensitive data, but is able to leverage prior knowledge (public, but possibly erroneous or untrusted) about the data distribution.

  We design private algorithms in this augmented setting for three flagship distribution testing tasks, \emph{uniformity}, \emph{identity}, and \emph{closeness} testing, whose sample complexity smoothly scales with the claimed quality of the auxiliary information. We complement our algorithms with information-theoretic lower bounds, showing that their sample complexity is optimal (up to logarithmic factors).
\end{abstract}

\vspace{-0.75em}\section{Introduction}\vspace{-0.25em}
Accurately analyzing data while preserving individual privacy is a fundamental challenge in statistical inference. 
Since its formulation nearly two decades ago, Differential Privacy (DP)~\citep{DworkMNS06} has emerged as the leading framework for privacy-preserving data analysis, providing strong mathematical privacy guarantees and gaining adoption by major entities such as the U.S. Census Bureau, Amazon~\citep{aws_clean_rooms_2024}, Google~\citep{erlingsson2014rappor}, Microsoft~\citep{ding2017collecting}, and Apple~\citep{apple2017learning, thakurta2017learning}.

Unfortunately, DP guarantees often come at the cost of increased data requirements or computational resources, which has limited the widespread adoption of differential privacy in spite of its theoretical appeal. To address this issue,
a recent line of work has investigated whether access to even small amounts of additional \emph{public} data could help mitigate this loss of performance.
Promising results for various tasks
have been shown, both experimentally~\citep{KST20, LowyLHR24, BuZHZK24, DaumORKSF24} and theoretically~\citep{Bie0S22, Ben-DavidBC0S23}. 
The use of additional auxiliary information
is very enticing, as such access
is available in many real-world applications: for example, hospitals handling sensitive patient data might leverage public datasets, records from different periods or locations, or synthetic data generated by machine learning models to improve analysis. Similarly, medical or socio-econonomic studies focusing on a minority or protected group can leverage statistical data from the overall population.

However, integrating public data introduces its own challenges, as it often lacks guarantees regarding its accuracy or relevance to private datasets. If the external data distribution deviates significantly from the target population, then it becomes at best useless for the task at hand, and at worst misleading; yet assessing its reliability prior to using it can be as complex as the original inference task. This leads to the following question, which is the focus of our work:\vspace{-0.25em}
\begin{framed}\itshape
\centering
\noindent{}How can we design private algorithms which
optimally utilize auxiliary information in
a way that seamlessly adapts to its (unknown, arbitrary) quality?
\end{framed}
We formalize the question by extending the recently proposed framework of ``augmented testing''~\citep{aliakbarpourIRS24} to the differentially private setting (see~\cref{sec:prelims} for formal definitions). 
In our new framework, 
the testing algorithm is provided with the auxiliary information as an ``advice'' probability distribution $\hat{p}$, purported to be a good approximation of the unknown data distribution $p$, along with a claimed accuracy $\accuracy$ of this approximation. 
The algorithm must correctly solve the inference task when the advice is indeed $\accuracy$-accurate; but is allowed to abort and return a ``failure'' symbol when it detects that the advice does not meet the claimed accuracy $\accuracy$. The name of the game is to achieve better data utilization as a function of this new accuracy parameter $\accuracy$ by leveraging the advice when it is good, yet not being fooled by it when it is bad enough to derail the algorithm.

Our new model must take into account the privacy constraints of the samples.  
While
there is no need to consider the 
privacy of sample data $\hat{p}$ that is already public,
we must ensure that any other sample data remains private,  even if our algorithm decides to abort due to bad advice.

Finally, for the purpose of this paper, we primarily focus on distribution testing (which can be viewed as a finite-sample, computational take on hypothesis testing), the cornerstone of many scientific endeavors. However, we believe that differentially  private inference using (unverified) auxiliary public data is a promising avenue for work beyond the specific tasks we consider here.

\subsection{Related work}
Distribution testing has been extensively studied since its first formulation in~\citet{GGR98,BatuFRSW00}; we only refer below to the papers most relevant to us, and refer the reader to the expositions, textbooks, and surveys~\citep{Rubinfeld12,GoldreichBook17,CanonneSurvey20,Canonne22} for a more extensive coverage.
\emph{Uniformity testing}, that is, the task of deciding if an unknown distribution over a given domain of size $n$ is uniform, or at distance at least $\eps$ from it (in total variation), 
was shown to have sample complexity $\Theta(\sqrt{n}/\eps^2)$ in~\citet{GR00, GR11,Paninski08,AcharyaDK15,DiakonikolasGPP18}. It was shown to be equivalent to the more general task of \emph{identity} testing, where the reference distribution is allowed to be arbitrary (instead of uniform) in~\cite{Diakonikolas:2016,Goldreich16}. The sample complexity of \emph{closeness} testing, where the algorithm has access to samples from two unknown distributions and must decide whether they are equal or at total variation distance at least $\eps$, was settled in a series of works~\citep{BatuFRSW00,Valiant11,ChanDVV14,DiakonikolasGKP21,CanonneS22}, culminating in the tight sample complexity $\Theta(\sqrt{n}/\eps^2+n^{2/3}/\eps^{4/3})$.

Beyond this standard distribution testing setting, the work closest to ours is the recent paper of~\citet{aliakbarpourIRS24}, which introduced the (non-private) framework of augmented testing for distributions. While there has been before a large body of work on learning-augmented algorithms (and previously, on algorithms with advice) (e.g.,~\citet{HsuIKV19,indyk2019learning,jiang2020learningaugmented,AamandCNSV23,BhattacharyyaCJG24}), the specific formulation whereby the algorithm is provided with advice in the form of a hypothesis probability distribution, \emph{and is allowed to abort if it detects this advice is incorrect}, is crucial to obtaining non-vacuous formulations of the problem. While~\citet{aliakbarpourIRS24} do address identity (and uniformity) as well as closeness testing of distributions in their augmented testing framework, they only consider the non-private versions of these tasks: introducing the constraint of differential privacy on the data (but not on the advice itself, considered public) is the main novelty of our work, and comes with a host of technical and conceptual challenges to overcome. 
We refer the reader to the website \url{https://algorithms-with-predictions.github.io/}, for more on learning-augmented algorithms.

On the privacy-preserving side, there is a large literature on DP distribution testing, where the samples provided to the algorithm are considered sensitive and the algorithm must satisfy the constraint of (central) DP, notably~\citet{CaiDK17,AcharyaSZ18,ADR18,ADKR19}. The tight private sample complexity of uniformity, identity,\footnote{Note that the equivalence between uniformity and identity testing of~\cite{Diakonikolas:2016,Goldreich16} carries over to the differentially private setting.} and closeness testing under $\privacyeps$-DP are now known to be
$
\Theta(\sqrt{n}/\eps^2+\sqrt{n}/(\eps\sqrt{\privacyeps}) + n^{1/3}/(\eps^{4/3}\privacyeps^{2/3}) + 1/(\eps\privacyeps))
$
and 
$
\Theta(\sqrt{n}/\eps^2 + n^{2/3}/\eps^{4/3}+\sqrt{n}/(\eps\sqrt{\privacyeps}) + n^{1/3}/(\eps^{4/3}\privacyeps^{2/3}) + 1/(\eps\privacyeps))
$, respectively~\citep{ZhangThesis21}.

As mentioned earlier, a recent line of work has investigated the use of additional public data for DP inference questions, and in particular for distribution learning questions~\citep{Bie0S22, Ben-DavidBC0S23}. We emphasize that while these results can accommodate a (small) amount of \emph{distribution shift} between the private and public data distributions, they assume that the publicly available data \emph{is} accurate (comes from the same distribution as, or one very close to, the the sensitive data). This is a crucial limiting assumption, and one our paper does not make: in our setting, we do not have to trust the quality of the auxiliary data. Our framework for distribution testing bears some resemblance to the framework of \cite{khodak2023} for learning-augmented private quantile estimation, in which the algorithm must be correct (and efficient) even if the prediction is poor.  

Another important difference is in the parameter regime:~\citep{Bie0S22, Ben-DavidBC0S23} consider the setting where the quantity of public data is small compared to the amount of public data: that is, they investigate whether the algorithm can benefit from a few ``useful'' public data points, in order to perform its analysis on a much larger sensitive dataset. In contrast, we focus on the setting where the public data abounds (so that even using it to get a full probability distribution $\hat{p}$ as advice is possible), but the private data is much scarcer.

\subsection{Our results}

Our main results are sample-efficient private algorithms for identity and closeness testing in the augmented setting, complemented by (nearly)-matching information-theoretic lower bounds.
\begin{theorem}[{Informal version of~\cref{thm:idupperbound} and~\cref{thm:idlowerbound}}]
    There is an algorithm for private augmented identity testing of distributions over $[n]$ which, given a reference distribution $q$, privacy parameter $\privacyeps>0$, distance parameter $\testingeps\in(0,1]$, purported accuracy $\accuracy\in(0,1]$, as well as advice distribution $\hat{p}$, takes
    \[
        \Theta \left( \frac{\sqrt{n}}{\testingeps^2} + \frac{\sqrt{n}}{\testingeps\sqrt{\privacyeps}} + \frac{n^{1/3}}{\testingeps^{4/3}\privacyeps^{2/3}} + \frac{1}{\testingeps\privacyeps} \right)
    \]
    samples when the distance $\eta \coloneqq \totalvardist{\hat{p}}{q}$ satisfies $\eta \leq \accuracy$, and
    \[
    \Theta \left( \min \left( \frac{1}{(\eta-\accuracy)^2} + \frac{1}{(\eta - \accuracy)\xi}, \frac{\sqrt{n}}{\testingeps^2} + \frac{\sqrt{n}}{\testingeps\sqrt{\privacyeps}} + \frac{n^{1/3}}{\testingeps^{4/3}\privacyeps^{2/3}} + \frac{1}{\testingeps\privacyeps} \right) \right)
    \]
    when $\eta > \accuracy$. Moreover, this is optimal.
\end{theorem}
Observe that in the ``interesting regime''~--~i.e., when the advice suggests to the algorithm that it should reject, as the claimed distance between $\hat{p}$ and the unknown distribution $p$ is strictly smaller than the distance between $\hat{p}$ and the reference distribution $q$~--~and, in particular, for vanishing $\accuracy$, the resulting sample complexity can become as small as $\Theta(1/\eta^2+1/(\eta\privacyeps))$, leading to stark savings in the amount of private data required to conduct the task.

Note that the above theorem also applies to uniformity testing in a straightforward manner. In particular, by setting $q$ as the uniform distribution, our identity testing algorithm immediately applies to uniformity testing, and our lower bound—proven under the assumption of uniform $q$—establishes the complete equivalence of these two problems.

\noindent Our results for the task of closeness testing are summarized in the next theorem:
\ifcolt
\begin{theorem}[Informal version of~\cref{thm:closenessupperbd} and~\cref{thm:closenesslowerbd}]
\else
\begin{theorem}[Informal version of~\cref{thm:closenessupperbd:restated} and~\cref{thm:closenesslowerbd}]
\fi
    \label{theo:informal:closeness}
    There is an algorithm for private augmented closeness testing of distributions over $[n]$ which, given privacy parameter $\privacyeps>0$, distance parameter $\testingeps\in(0,1]$, purported accuracy $\accuracy\in(0,1]$, as well as advice distribution $\hat{p}$ for one of the two distributions, takes
    \begin{equation*}
        \tilde{O} \!\left( \frac{n^{2/3}\accuracy^{1/3}}{\testingeps^{4/3}} + \frac{\sqrt{n}}{\testingeps^2} + \frac{n^{1/3}}{\testingeps^{4/3}\privacyeps^{2/3}} + \frac{\sqrt{n}}{\testingeps\sqrt{\privacyeps}} + \frac{1}{\testingeps\privacyeps} \right) 
    \end{equation*}
    samples. Moreover, this is optimal up to logarithmic factors in $n$.
\end{theorem}
Again, observe that the advice influences the first term of the sample complexity. As the claimed advice quality $\accuracy$ goes to 0 (and ignoring logarithmic factors), our results show that, surprisingly, the task becomes essentially as simple as differentially private \emph{identity} testing.

\subsection{Overview of our technical contributions}
\myparagraph{Upper bounds.} The {\em flattening technique}, introduced in~\cite{DiakonikolasK16}, is an important tool in distribution testing that has found applications in various problems, including closeness testing.
This technique maps a distribution \( p \) to a ``flattened distribution'' \( p' \) with a lower \(\ell_2\)-norm. 
This reduces the testing task (with respect to total variation distance) to the much more manageable analogous task with respect to $\ell_2$ distance, while controlling the blowup in sample complexity. At a high level, the technique breaks any high-probability element into multiple new domain elements (buckets) by assigning its probability mass equally among them. The key problem is to determine the number of buckets required for each element \( i \), which can vary depending on the problem's structure.

One common approach proposed in~\cite{DiakonikolasK16} is as follows. Suppose we have \( k \) samples from an unknown distribution \( p \), referred to as {\em flattening samples}. Let \( k_i \) denote the frequency of element \( i \) among these samples. For each element \( i \), we assign \( k_i + 1 \) buckets. Elements which appear more frequently (i.e., the large elements) are split into more buckets, reducing their individual probability mass. More formally, one can show that the resulting distribution $p'$ has an \(\ell_2\)-norm of at most \( 1/k \) in expectation. Note that we can simulate a sample from \( p' \) by drawing a fresh sample \( i \) from \( p \) and assigning it to a random bucket in \([k_i + 1]\).

Why is this result significant for closeness testing (or related problems)? One can show that if the same flattening is applied to two distributions, their \(\ell_1\)-distance remains unchanged. Thus, testing the closeness of \( p \) and \( q \) reduces to testing the closeness of their flattened versions \( p' \) and \( q' \). Since the flattened distributions have a reduced \(\ell_2\)-norm, we can leverage the \(\ell_2\)-tester from~\citet{ChanDVV14}, whose sample complexity is proportional to the \(\ell_2\)-norm, thereby achieving more sample-efficient testing. 
This technique is particularly appealing because it allows us to exploit problem-specific structure to design an effective flattening scheme. For example, Aliakbarpour et al.~\citep{aliakbarpourIRS24} use a prediction distribution to guide the flattening: if the prediction assigns a higher probability to an element, more buckets are allocated to it, further reducing its \(\ell_2\)-contribution. After flattening, they test whether the \(\ell_2\)-norm of the distribution is sufficiently small. If not, they conclude that the prediction quality was poor. If the norm is small, the \(\ell_2\)-tester from~\cite{ChanDVV14} can be used with significantly fewer samples than would be needed without the prediction, yielding an augmented closeness tester with an optimal dependence on the prediction quality.

To achieve our goals, a natural approach is thus to privatize the algorithm from~\cite{aliakbarpourIRS24}. Unfortunately, the sample-based flattening step introduces privacy challenges, because flattening via samples is highly sensitive: changing a single sample can alter the number of buckets for two elements, drastically affecting the structure of the resulting distribution. To ensure privacy, the outcomes of these very different cases must be made similar, typically by adding a large amount of noise. Consequently, statistics computed on the flattened distribution would require a large number of samples to be privatized, leading to suboptimal algorithms.

To circumvent this issue, we need to reduce the sensitivity of the flattening step. In order to do so, we leverage the framework of~\cite{ADKR19}, which focuses on privatizing a flattening-based tester~--~exactly what we need for our approach to work! However, this is not as straightforward as it seems.
Because the advice distribution $\hatp$ used as a guide in the flattening step may be wildly inaccurate, we must verify that the outcome of the flattening step is satisfactory; that is, that the $\ell_2$ norm of the flattened distribution is sufficiently small (as it should be, were the advice correct). To further complicate matters, this verification must be done \emph{privately}. 

Our solution is to apply flattening in two steps. First, we use the prediction $\hatp$ to flatten $p$ into \( p' \). Then, we use samples drawn from \( p \) (more precisely, from \( p' \)) to further flatten the distribution into \( p'' \). If \( \hat{p} \) is \(\accuracy\)-close to the true distribution, we expect that after these two steps the \(\ell_2\)-norm of \( p'' \) is roughly less than \(\accuracy/k\). We separate these two steps due to our privacy requirements; since the prediction is public information, there is no privacy concern in the first step, which simplifies the analysis of the second step.

The standard statistic for computing the \(\ell_2\)-norm is the number of collisions in a sample set (i.e., the number of pairs of equal samples)~\citep{GR11}. This statistic is highly sensitive to changes in one sample. For example, consider estimating the \(\ell_2\)-norm using \(\ell\) samples. Suppose element \( i \) appears \(\ell/2\) times. If we assign one bucket to \( i \), the number of collisions is \({\ell/2 \choose 2}\). However, if we assign two buckets to \( i \), we expect each bucket to receive roughly \(\ell/4\) samples, so the total number of collisions is \(2{\ell/4 \choose 2}\). This change reduces the number of collisions by roughly \(\ell^2/16\), whereas ideally we would like this number to be a constant.

Thankfully, this worst-case only occurs in highly atypical scenarios. In general, if we draw \( k \) samples for flattening, we expect about \( kp_i \) buckets for element \( i \). Then if we draw \(\ell\) samples for estimation, we expect approximately \( \ell p_i \) samples from \( i \). Distributing these equally among the buckets yields roughly \(\ell/k\) samples per bucket. Assuming \( k = \ell \), a typical dataset should have a constant number of samples per bucket. Consequently, altering either a flattening sample or an estimation sample would change the overall number of collisions by at most a constant factor. Hence, the sensitivity of the statistic should be low for a ``typical'' dataset.

The challenge is thus to pay only a privacy cost proportional to the sensitivity of a typical dataset, rather than the worst-case dataset. To this end, we identify the ``bad events'' that cause our dataset to have high sensitivity, which correspond to having a high unbalance between samples either within buckets, or between flattening and estimation sets. That is, we must ensure that the following holds for every element $i \in [n]$:
\begin{enumerate}
\setlength{\itemsep}{1pt}
\setlength{\parskip}{0pt}
\setlength{\parsep}{0pt}
    \item When the samples of $i$ samples are distributed among buckets, no bucket receives an unreasonably high number of samples.
    \item The number of instances of $i$ in the estimation sample set is not much larger than the number of instances of $i$ in the flattening sample set.
\end{enumerate}

To address the first issue, we mitigate the randomness inherent in distributing samples among buckets. The original statistic sums the collisions across buckets, where if \(\ell_{i,j}\) denotes the number of samples in the \(j\)-th bucket of element \( i \), the contribution is \({\ell_{i,j} \choose 2}\). We replace this quantity with its expectation under the assumption of a uniform distribution of samples among buckets.

The second issue is more complex. We need to ensure that the number of instances of element \( i \) is approximately the same in both the flattening and estimation sample sets. To achieve this, we employ a technique from~\cite{ADKR19}. Roughly speaking, each high-sensitivity sample set \( X \) is randomly mapped to another sample set \( X' \) such that any imbalance in the number of samples for an element is corrected. The mapping has the key property of preserving the Hamming distance between datasets. More precisely, if two datasets \( X_1 \) and \( X_2 \) (differing in one element) are mapped to low-sensitivity datasets \( Y_1 \) and \( Y_2 \), then there exists a coupling between \( Y_1 \) and \( Y_2 \) such that their Hamming distance is bounded by a constant. This property ensures that applying a private algorithm to the \( Y \)'s (with low sensitivity) yields privacy guarantees for the original \( X \)'s.

In our context, we ensure that the transformed dataset has a balanced number of instances for each element across both the estimation and flattening sets. We then run our private algorithm on this new sample set. If the original sample set already had low sensitivity, it remains unchanged, and we obtain the correct answer with high probability. If the dataset had high sensitivity, it is transformed to ensure privacy; although this may render the statistical result unreliable, such high-sensitivity datasets occur with low probability.\footnote{For readers familiar with differential privacy, note that our approach differs from the well-known ``Propose-Test-Release'' mechanism, where noise is added proportional to the typical sensitivity (see Section~3 in~\cite{vadhan2017complexity}). In that mechanism, the probability of a high-sensitivity (bad) event is absorbed into an approximate DP guarantee via an additive error \(\delta\). However, our approach is more suitable for pure DP: our privacy guarantee remains intact, and the bad event only introduces a small probability of error in accuracy.}

With these technical components, we obtain a low-sensitivity statistic for the \(\ell_2\)-norm and apply the Laplace mechanism to ensure privacy. Plugging this into the above outline yields our main algorithmic result. 
\ifcolt
A complete proof is given in Appendix~\ref{sec:UB_closeness}, with a short version is presented in \Cref{sec:UB_closeness_main_body}. 
\else
A complete proof is given in \Cref{sec:UB_closeness}. 
\fi

The upper bound for identity and uniformity testing, can be obtained via simple prioritization of the upper bound in~\cite{aliakbarpourIRS24} and the existing private algorithm for identity testing~\cite{AcharyaSZ18}.

\myparagraph{Lower bounds.}
For the lower bounds, the difficulties are reversed. 
The lower bound for private augmented closeness testing is derived by combining two existing lower bounds; further details are provided in~\cref{sec:LB_closeness}.

For uniformity testing, and consequently for identity testing, 
to show a lower bound, we must combine the statistical inference and privacy constraints on a private augmented tester. When $\hatp$ and $q$ are closer than the suggested accuracy $\accuracy$, we show that private identity testing reduces to augmented private identity testing using a similar reduction to \citet{aliakbarpourIRS24}. In this case, known lower bounds \citep{ZhangThesis21} automatically apply to our setting. 

However, if $\hatp$ and $q$ are more than $\accuracy$-far apart, we must combine several tools for proving information-theoretic lower bounds. In the non-augmented setting, a lower bound can be shown using a version of Le Cam's method for differential privacy \citep{AcharyaSZ18}. Informally, the idea is to construct a coupling between two distributions (one close to the uniform distribution, one far) whose expected Hamming distance is low given too few samples. Since no answer will succeed for both, and no private algorithm can answer too differently on the two sets of samples, no algorithm can succeed with high probability. In our case, there are \emph{three} possible answers, which requires us to construct three distributions such that no single answer works for all three. It also does not suffice to directly follow the approach of \citet{aliakbarpourIRS24} and construct a multivariate coupling between the three distributions such that all three are close in Hamming distance. This would prove the privacy terms of our lower bound, but would not give the correct combination of statistical and privacy terms. 

To avoid this problem, we use a general version of Le Cam's method under privacy and statistical constraints. Informally, our result states that if any algorithm requires $s_1$ samples to statistically distinguish two distributions with high probability and $s_2$ samples to guarantee that they are far in Hamming distance, it requires $\max(s_1,s_2)$ samples to distinguish them privately. We then construct a hard distribution which is $\testingeps$-far from the uniform distribution, and a hard distribution which is $\accuracy$-close to the prediction $\hatp$. Any private augmented uniformity tester cannot answer similarly for all three distributions, so it must distinguish the uniform distribution from at least one of the hard distributions. Therefore, the sample complexity must satisfy at least one of the constraints imposed by our version of Le Cam's method. This explains why the optimal sample complexity is the minimum of two expressions, which matches our upper bound. For more details, see \cref{sec:LB_uniformity}. \vspace{-1em}

\section{Preliminaries and Problem Setup}
    \label{sec:prelims}
 We use $[n]$ to indicate the set of integers $\{1,2,\dots, n\}$, and standard asymptotic notation ($O(\cdot), \Omega(\cdot), \Theta(\cdot)$) as well as the (slightly) less standard $\tilde{O}(\cdot)$, which omits polylogarithmic factors in its argument. In this paper, we focus on discrete probability distributions with a finite domain, which we identity with their probability mass functions (pmf). Let $p$ denote a probability distribution over $[n]$: for any $i\in[n]$, we denote by $p_i$ the probability of the element $i$ according to $p$.  
For a subset $S\subset[n]$, $p(S)$ then denotes the probability of observing an element in $S$ according to $p$: $p(S) = \sum_{i\in S} p_i$. The distribution of $s$ i.i.d. samples from $p$ is written as $p^{\otimes s}$. If we know the probability $p_i$ of each element, we say $p$ is a \textit{known distribution}. %
We also use the following standard probability distributions: $\Poi{\lambda}$ denotes a Poisson distribution with mean $\lambda$; $\Bern{x}$ denotes a Bernoulli distribution with success probability $x$; and $\Lap{b}$ denotes a Laplace distribution with scale $b$. 

The total variation distance between $p$ and $q$ is defined as the maximum possible difference in probabilities assigned to any event (subset of outcomes):
\[
\totalvardist{p}{q} \coloneqq \max_{S \subseteq [n]} \abs{p(S) - q(S)} = \frac{1}{2}\|p-q\|_1 \in [0,1].
\]
where for the last expression we identify the pmf of $p,q$ with their vector of probabilities. 
For two distributions $p$ and $q$, and any $\testingeps \in [0,1]$, we say $p$ is $\testingeps$-close to $q$ iff $\totalvardist{p}{q}$ is at most $\testingeps$. We say $p$ and $q$ are $\testingeps$-far from each other iff $\totalvardist{p}{q}$ is greater than $\testingeps$. A property $\PP$ is a set of distributions. We say a distribution $p$ has property $\PP$ if $p \in \PP$, and $p$ is $\testingeps$-far from property $\PP$ if $\totalvardist{p}{q} > \testingeps$ for all $q \in \PP$.

\myparagraph{Problem setup:} The standard distribution testing problem is to decide, given samples from an unknown distribution $p$, whether $p$ has a property $\PP$ or whether $p$ is $\testingeps$-far from property $\PP$. In our case, the tester receives three additional inputs: a privacy parameter $\privacyeps$, which it must respect, a predicted distribution (or ``advice") $\hatp$, and a suggested accuracy value $\accuracy$. If $\hatp$ is $\accuracy$-close to $p$, the tester must either output $\accept$ or $\reject$ with high probability; otherwise, it may output $\invalid$, which denotes inaccurate information.

\begin{definition}[Augmented private tester] Suppose we are given three parameters $\testingeps, \accuracy\in (0,1)$, and $\privacyeps>0$, along with distributions $p$ and $\hatp$ over the same domain. Suppose
$\AA$ receives samples from $p$ and returns a value in $\{\accept, \reject, \invalid\}$.
We say algorithm $\AA$ is a \textit{$(\privacyeps, \testingeps, \accuracy, \errprob)$-private augmented tester} for property $\PP$ iff the following holds:
\begin{itemize}
\setlength{\itemsep}{1pt}
\setlength{\parskip}{0pt}
\setlength{\parsep}{0pt}
    \item $\AA$ is a $\privacyeps$-differentially private algorithm. 
    \item If $p \in \PP$, $\AA$  does not output $\reject$ with probability more than $\errprob$. 

    \item If $p$ is $\testingeps$-far from $\PP$, $\AA$ does not output $\accept$ with probability more than $\errprob$. 

    \item If $p$ and $\hatp$ are $\accuracy$-close, $\AA$ does not output $\invalid$ with probability more than $\errprob$.  
\end{itemize}

\end{definition}
\begin{remark}[On knowing the parameter $\alpha$]
While it may seem difficult to suggest an accuracy level $\accuracy$ for an unverified prediction, this is not an actual limitation, as  one can avoid this in practice using the search algorithm of \cite{aliakbarpourIRS24}. The algorithm iteratively increases $\accuracy$ until the tester either returns $\accept$ or $\reject$. If the true accuracy is $\accuracy^*$ and the tester uses $f(\accuracy)$ samples for a single run with input $\accuracy$, the search algorithm uses $O(f(\accuracy^*))$ total samples in expectation.

While we do not provide the full proof here, careful analysis shows that a private version of this claim can be established under certain constraints. The main challenge is determining the stopping point at which the tester privately returns either \(\accept\) or \(\reject\). This can be accomplished using the well-known Sparse Vector Technique, which does not increase the privacy cost by more than a constant factor, provided that the stopping point can be replaced with a threshold query to a low-sensitivity function. Notably, this condition holds for all the algorithms in this paper.
\end{remark}

We will refer to a non-private augmented tester as a $(\testingeps, \accuracy, \errprob)$-augmented tester. We will refer to a non-augmented private tester as a $(\privacyeps, \accuracy, \errprob)$-private tester. Finally, a $(\testingeps, \errprob)$-tester is simply a standard, non-private, non-augmented tester. 

We focus on two common variants of the distribution testing problem in which $\PP$ contains a single distribution $q$, and the problem is to distinguish between the cases $p=q$ and $\totalvardist{p}{q} > \testingeps$. In the case of \textit{identity testing}, $q$ is a known distribution. In the case of \textit{closeness testing}, $q$ is unknown, and the algorithm receives samples from both $p$ and $q$.

\myparagraph{Differential privacy:} For two datasets $X$ and $X'$, we write the Hamming distance of $X$ and $X'$ -- the number of samples in which the two datasets differ -- as $\Ham{X}{X'}$. We consider an algorithm to be private if it satisfies the following definition.

\begin{definition}[Differential privacy]
    Fix parameter $\privacyeps > 0$. Let $X, X' \in \mathcal{X}$ be two datasets with $\Ham{X}{X'} = 1$; that is, $X$ and $X'$ differ in exactly one sample. A randomized algorithm $\AA: \mathcal{X} \rightarrow \{\accept, \reject, \invalid \}$ is \textit{$\privacyeps$-differentially private} if for all such $X$ and $X'$ and any $O \subseteq \{\accept, \reject, \invalid \}$,
    $
        \Pr{\AA(X) \in O} \leq e^{\privacyeps} \cdot \Pr{\AA(X') \in O}\,.
    $
\end{definition}

Our algorithms use the Laplace mechanism~\citep{DworkR14}, which works as follows. Let $\mathcal{X}$ be the set of all datasets and $f\colon \mathcal{X} \rightarrow \mathbb{R}$ be a real-valued function. Define the sensitivity of $f$, denoted $\Delta(f)$, as
$
    \Delta(f) \coloneqq \max_{X, X' \in \mathcal{X} : \Ham{X}{X'} = 1} |f(X) - f(X')|\,. 
$
The quantity $f(X) + \Lap{\Delta(f)/\privacyeps}$ is then $\privacyeps$-differentially private. 

In this paper, we focus on pure differential privacy in the central setting, and implicitly focus on the ``high-privacy regime'' (range $\privacyeps\in(0,1]$). We leave other variants (e.g., local DP or approximate DP) as open problems. 

\myparagraph{Organization of the paper:} The upper bound for identity testing (and, equivalently, uniformity testing) is presented in \Cref{sec:UB_identity}, while the lower bound for uniformity testing (and hence identity testing) appears in \Cref{sec:LB_uniformity}. The upper bound for closeness testing is given in \Cref{sec:UB_closeness} and the lower bound is in \Cref{sec:LB_closeness}. 

\section{Upper bound for Private Augmented Identity Testing}
\label{sec:UB_identity}

Our upper bound follows directly from the upper bound of \cite{aliakbarpourIRS24} for non-private augmented identity testing. The tester aims to detect the case where $p$ is far from $q$ using the Scheffé set $S$ of $\hat{p}$ and $q$, where $S \coloneqq \{i \in [n]: \hat{p}_i < q_i \}$. The difference in the probability of $S$ under $\hat{p}$ and $q$ is precisely the total variation distance:
\begin{equation*}
    \eta \coloneqq \totalvardist{q}{\hat{p}} = |q(S) - \hat{p}(S)|
\end{equation*}
The algorithm draws samples from $p$ and computes the test statistic $\sigma$, the fraction of samples which fall in $S$. If $\sigma$ is sufficiently far from $q(S)$, the tester rejects; otherwise, it returns $\invalid$. The full pseudocode of our private tester is given in Algorithm~\ref{alg:idtester}. 

The sample complexity of the non-private augmented tester depends on $\eta$ and $\accuracy$:
\begin{lemma}[\cite{aliakbarpourIRS24}]
    Let $\testingeps \in (0,1], \accuracy \in [0,1)$. Suppose we are given known distributions $q$ and $\hat{p}$, and unknown distribution $p$, all over $[n]$. Let $\eta \coloneqq \totalvardist{q}{\hat{p}}$ and $\errprob_0\coloneqq 1/10$. There is an $(\testingeps, \accuracy, \errprob_0)$-augmented identity testing algorithm which takes $s$ samples from $p$, where 
    \begin{equation*}
    s = \begin{cases} \Theta\left(\frac{\sqrt{n}}{\testingeps}\right) & \text{if $\eta \leq \accuracy$} \\
    \Theta \left( \min \left(\frac{1}{(\eta-\accuracy)^2}, \frac{\sqrt{n}}{\testingeps^2} \right) \right) & \text{if $\eta > \accuracy$}
    \end{cases}
\end{equation*}
\end{lemma}
This result shows an improvement in sample complexity if $\hat{p}$ is a sufficiently good estimate of $p$, and does no worse than a standard identity tester if not. To privatize the tester, we simply add Laplace noise to the test statistic $\sigma$. 

\begin{lemma}
    \label{lem:idprivacy}
    Fix privacy parameter $\privacyeps > 0$. Let $\hat{p}$ and $q$ be known distributions, and $p$ be an unknown distribution, over $[n]$. Let $S \coloneqq \{i \in [n]: \hat{p}_i < q_i \}$. For a set $\{x_1,...,x_s\}$ of $s$ i.i.d. samples from $p$, let  
    \begin{equation*}
        \sigma \coloneqq \frac{1}{s} \sum_{j=1}^s \mathds{1} \left[ x_j \in S \right] 
    \end{equation*}
    Then the statistic $\hat{\sigma} = \sigma + \Lap{1/(s\privacyeps)}$ is $\xi$-differentially private.
\end{lemma}
\begin{proof}
    The sensitivity of $\sigma$ is
    \begin{equation*}
        \Delta(\sigma) = \max_{X, X': |X - X'| < 1} |\sigma(X) - \sigma(X')| = \frac{1}{s} 
    \end{equation*}
    Therefore, adding $\Lap{1/(s\privacyeps)}$ noise suffices to make the statistic private by the Laplace mechanism. 
\end{proof}

In the case where augmentation does not decrease the sample complexity, we must run a non-augmented private identity tester. The following lemma gives the sample complexity of private identity testing.
\begin{lemma}[\cite{AcharyaSZ18}]
    \label{lemma:privateidtesting}
    Fix parameters $\testingeps \in (0,1]$ and $\privacyeps > 0$. Let $q$ be a known distribution and $p$ be an unknown distribution over $[n]$. There is a $(\privacyeps, \testingeps, \delta=0.1)$-private identity tester which takes $s$ samples from $p$, where
    \begin{equation}
        s = \Theta \left( \frac{\sqrt{n}}{\testingeps^2} + \frac{\sqrt{n}}{\testingeps\sqrt{\privacyeps}} + \frac{n^{1/3}}{\testingeps^{4/3}\privacyeps^{2/3}} + \frac{1}{\testingeps\privacyeps} \right)
    \end{equation}
    Additionally, any $(\privacyeps, \testingeps,\delta=0.1)$-private identity tester requires $\Omega(s)$ samples.
\end{lemma}

\begin{algorithm}[t]
\setcounter{AlgoLine}{0}
\caption{\textsc{Private-Augmented-Identity-Tester}}\label{alg:idtester}
\KwIn{$p, q, \hat{p}, n, \testingeps, \accuracy, \privacyeps, \errprob=0.1$}
\KwOut{$\accept$, $\reject$, or $\invalid$} 

\DontPrintSemicolon
\nl $\eta \gets \totalvardist{\hat{p}}{q}$ \; 

\nl \uIf{$\eta \leq \accuracy$ or $\frac{\sqrt{n}}{\testingeps^2} + \frac{\sqrt{n}}{\testingeps\sqrt{\privacyeps}} + \frac{n^{1/3}}{\testingeps^{4/3}\privacyeps^{2/3}} + \frac{1}{\testingeps\privacyeps} \leq \frac{1}{(\eta-\alpha)^2} + \frac{1}{(\eta - \accuracy)\xi}$} {
    \nl Run the tester of Lemma~\ref{lemma:privateidtesting} and output its answer \;
}
\nl $S \gets \{i \in [n] : \hat{p}_i < q_i \}$ \;

\nl Draw $s = \Theta \left( \frac{1}{(\eta-\alpha)^2} + \frac{1}{(\eta - \accuracy)\xi} \right)$ samples $x_1,...x_s$ from $p$ \;

\nl $\sigma \gets \frac{1}{s} \sum_{j=1}^n \mathds{1}[x_j \in S]$ \;

\nl $\hat{\sigma} \gets \sigma + \Lap{\frac{1}{s\privacyeps}}$ \;

\nl \uIf {$|\hat{\sigma} - q(S)| > \frac{\eta-\accuracy}{4}$} {
    \nl \Return $\reject$ \;
} 
\nl \uElse {
    \nl \Return $\invalid$ \;
}
\end{algorithm}

Combining the results above, we have the following upper bound.

\begin{theorem}
\label{thm:idupperbound}
Fix parameters $\testingeps \in (0,1]$, $\accuracy \in [0,1)$, $\xi > 0$, and $\delta_0 \coloneqq 1/10$. Let $\hat{p}$ and $q$ be known distributions, and $p$ be an unknown distribution, over $[n]$. Then Algorithm~\ref{alg:idtester} is a $(\privacyeps, \testingeps, \accuracy, \delta_0)$-private augmented identity tester which takes $s$ samples from $p$, where 
\begin{align}
    s = \begin{cases}
        \Theta \left( \frac{\sqrt{n}}{\testingeps^2} + \frac{\sqrt{n}}{\testingeps\sqrt{\privacyeps}} + \frac{n^{1/3}}{\testingeps^{4/3}\privacyeps^{2/3}} + \frac{1}{\testingeps\privacyeps} \right), & \text{$\eta \leq \alpha$} \\
        \Theta \left( \min \left( \frac{1}{(\eta-\alpha)^2} + \frac{1}{(\eta - \accuracy)\xi}, \frac{\sqrt{n}}{\testingeps^2} + \frac{\sqrt{n}}{\testingeps\sqrt{\privacyeps}} + \frac{n^{1/3}}{\testingeps^{4/3}\privacyeps^{2/3}} + \frac{1}{\testingeps\privacyeps} \right) \right), & \text{$\eta > \alpha$} 
    \end{cases}
\end{align}
\end{theorem}
\begin{proof}
Note that the choice to use one algorithm or the other only depends on the (public) parameters, not on the samples: therefore, the decision of which algorithm to use does not compromise privacy. 

    If $\eta \leq \accuracy$ or $\sqrt{n}/\testingeps^2 + \sqrt{n}/(\testingeps\sqrt{\privacyeps}) + n^{1/3}/(\testingeps^{4/3}\privacyeps^{2/3}) + 1/(\testingeps\privacyeps) \leq 1/(\eta-\accuracy)^2 + 1/(\xi(\eta-\accuracy))$, we run the non-augmented private identity tester, which is $\privacyeps$-differentially private and returns the correct answer with probability at least $0.9$. The sample complexity in this case is given by Lemma~\ref{lemma:privateidtesting}. 

    Next, we consider the case where $\eta > \accuracy$ and $1/(\eta-\accuracy)^2 + 1/(\xi(\eta-\accuracy)) < \sqrt{n}/\testingeps^2 + \sqrt{n}/(\testingeps\sqrt{\privacyeps}) + n^{1/3}/(\testingeps^{4/3}\privacyeps^{2/3}) + 1/(\testingeps\privacyeps)$. We will separately prove privacy, correctness, and the sample complexity. 
    
    \paragraph{Privacy:} In the case where $\eta \leq \accuracy$ or the non-augmented tester has lower sample complexity than the augmented version, we run the non-augmented private identity tester, which is $\privacyeps$-differentially private by Lemma~\ref{lemma:privateidtesting}. If we run the augmented tester, the algorithm is $\privacyeps$-differentially private by Lemma~\ref{lem:idprivacy}. 
    
    \paragraph{Correctness:} Note that the augmented tester can never return $\accept$. We will only consider the other two possibilities. First, we show that with high probability, the algorithm does not return $\reject$ when $p=q$. If $s = \Omega(1/(\eta-\alpha)^2)$, then by Hoeffding's inequality, 
    \begin{equation}
        \label{eq:hoeffding}
        \Pr{|\sigma - p(S)| \geq \frac{\eta-\accuracy}{8} } \leq 2\exp \left( -2\left( \frac{\eta-\alpha}{8} \right)^2 \cdot s \right) \leq 0.05
    \end{equation}
    By the cdf of the Laplace distribution, if $s = \Omega(1/((\eta - \accuracy)\xi))$,
    \begin{equation}
        \label{eq:laplaceconcentration}
        \Pr {|\hat{\sigma} - \sigma| \geq \frac{\eta-\accuracy}{8} } 
        = \exp \left( \frac{-(\eta-\accuracy)\xi}{8\Delta(\sigma)} \right) 
        = \exp \left( \frac{-(\eta-\accuracy)\xi \cdot s}{8} \right) \leq 0.05
    \end{equation}
    Combining ~\cref{eq:hoeffding} and ~\cref{eq:laplaceconcentration} and taking a union bound, the following holds with probability at least $0.9$:
    \begin{equation*}
        |\hat{\sigma} - p(S)| \leq |\hat{\sigma} - \sigma| + |\sigma - p(S)| \leq \frac{\eta-\alpha}{4}
    \end{equation*}
    If $p=q$, then $p(S) = q(S)$, and the algorithm returns $\reject$ with probability at most $0.1$. 

    Finally, we show that if $\totalvardist{p}{\hat{p}} \leq \accuracy$, then the algorithm does not return $\invalid$ with high probability. In this case, with probability at least $0.9$,
    \begin{align*}
        \alpha &\geq \totalvardist{p}{\hat{p}} \geq |p(S) - \hat{p}(S)| \\
        &\geq |p(S) - q(S)| - |q(S) - \hat{p}(S)| \\
        &= \totalvardist{p}{q} - |q(S) - \hat{p}(S)| \\
        &= \eta - |q(S) - \hat{p}(S)| \\
        &\geq \eta - |\hat{\sigma} - p(S)| - |\hat{\sigma} - q(S)| \\
        &\geq \eta - \frac{\eta - \accuracy}{4} - |\hat{\sigma} - q(S)| \\
    \end{align*}
    Therefore, 
    \begin{equation*}
        |\hat{\sigma} - q(S)| \geq \frac{3(\eta-\accuracy)}{4}
    \end{equation*}
    and the probability that the algorithm returns $\invalid$ is at most $0.1$. \\

    \paragraph{Sample Complexity:} We have used the fact that $s = \Omega(1/(\eta-\alpha)^2)$ and $s = \Omega(1/((\eta - \accuracy)\xi))$. Therefore, it suffices to have
    \begin{equation*}
        s = \Theta \left( \frac{1}{(\eta-\alpha)^2} + \frac{1}{(\eta - \accuracy)\xi} \right) 
    \end{equation*}
    which completes the proof. 
\end{proof}

\section{Lower bound for Private Uniformity Testing}
\label{sec:LB_uniformity}

Our lower bound can be split into two parts based on the regime of parameters. When $\accuracy \geq \eta$, we cannot improve on private identity testing, and we obtain our lower bound from known results. When $\accuracy < \eta$, augmented private testing may be able to achieve a lower sample complexity. In this case, we prove a lower bound using a version of Le Cam's method.  

\subsection{Lower Bound When $\accuracy \geq \eta$}

When the predicted distribution $\hatp$ is further from the known distribution $q$ than the desired accuracy (that is, $\eta \leq \accuracy$), standard identity testing can be reduced to augmented identity testing. It immediately follows from this that private identity testing can be reduced to private augmented identity testing. Therefore, the lower bound when $\eta \leq \accuracy$ is precisely the lower bound for private identity testing. We formalize this argument by combining two known results. 

\begin{lemma}[\cite{aliakbarpourIRS24}]
    \label{lem:reduction}
    Fix testing parameters $\testingeps, \accuracy \in [0,1/2)$ and privacy parameter $\privacyeps > 0$. Let $\hatp$ and $q$ be known distributions over $[n]$, and $p$ be an unknown distribution over $[n]$. Let $\eta \coloneqq \totalvardist{q}{\hatp}$. Then if there exists an $(\testingeps, \accuracy, \delta = 1/3)$-augmented identity tester which takes $s$ samples, there exists an $(\testingeps, \delta=1/3)$ standard identity tester which takes $s$ samples.  
\end{lemma}

\begin{theorem}
    Fix testing parameters $\testingeps, \accuracy \in [0,1/2)$ and privacy parameter $\privacyeps > 0$. Let $\hatp$ and $q$ be known distributions over $[n]$, and $p$ be a known distribution over $[n]$. Let $\eta \coloneqq \totalvardist{q}{\hatp}$. Then if $\eta < \accuracy$, any $(\privacyeps, \accuracy, \testingeps, \errprob = 0.2)$-private augmented identity tester for testing identity of $p$ and $q$ with prediction $\hatp$ requires $s$ samples, where
    \begin{equation}
        s \geq \Omega \left( \frac{\sqrt{n}}{\testingeps^2} + \frac{n^{1/3}}{\testingeps^{4/3}\privacyeps^{2/3}} + \frac{\sqrt{n}}{\testingeps\sqrt{\privacyeps}} + \frac{1}{\testingeps\privacyeps} \right) 
    \end{equation}
\end{theorem}
\begin{proof}
    By Lemma~\ref{lem:reduction}, identity testing can be reduced to augmented identity testing when $\eta \leq \accuracy$. Since the privacy constraint does not reduce the sample complexity necessary for correctness, private identity testing can be reduced to private augmented identity testing. Therefore, the lower bound for private identity testing given in Lemma~\ref{lemma:privateidtesting} is also a lower bound for augmented testing in this case. 
\end{proof}

\subsection{Lower Bound When $\accuracy < \eta$}

When $\accuracy < \eta$, it is sometimes possible for the augmented private tester to achieve a lower sample complexity than the non-augmented tester. To account for this possibility, our lower bound uses an extension of Le Cam's method to the augmented setting. At a high level, the idea is to construct three distributions such that no single answer is valid for all three. We show that, with too few samples, (1) the distributions are statistically indistinguishable from the uniform distribution and (2) under differential privacy constraints, no algorithm can answer differently for all three distributions with high probability. We conclude that any algorithm which uses too few samples must have high error probability on at least one distribution. Formally, our result is the following:

\begin{theorem}
    \label{thm:idlowerbound}
    Fix testing parameters $\testingeps, \accuracy \in [0,1/2)$ and privacy parameter $\privacyeps > 0$. Let $\hatp$ and $q$ be known distributions, and $p$ be an unknown distribution over $[n]$. Let $\eta \coloneqq \totalvardist{q}{\hatp}$. Then if $\eta > \accuracy$, any $\xi$-differentially private $(\accuracy, \testingeps, \errprob = 0.2)$-augmented testing algorithm for testing identity of $p$ and $q$ with prediction $\hatp$ requires $s$ samples, where
    \begin{align}
        \label{eq:idlowerbound}
        s \geq \Omega \left( \min \left( \frac{1}{(\eta - \accuracy)^2} + \frac{1}{\privacyeps(\eta - \accuracy) }, \frac{\sqrt{n}}{\testingeps^2} + \frac{n^{1/3}}{\testingeps^{4/3}\privacyeps^{2/3}} + \frac{\sqrt{n}}{\testingeps\sqrt{\privacyeps}} + \frac{1}{\testingeps\privacyeps} \right) \right)
    \end{align}
\end{theorem}

We defer the proof to the end of this section. We begin by establishing a general tool to combine statistical and privacy constraints. In particular, we show that if two sets of samples look similar with respect to either statistical indistinguishability or Hamming distance, any algorithm must behave similarly on those inputs. 

\begin{lemma}\label{lem:pairwise_indistinguishability}
Suppose we are given three positive integers $n$, $s_1$, and $s_2$, along with parameters $\privacyeps> 0$ and $\tau \in (0,1)$. Consider two distributions (or families of distributions) $p_1$ and $p_2$ over $[n]$. Let $T_1^{s}$ and $T_2^s$ indicate two sample set of size $s$ from $p_1$ and $p_2$ respectively. Assume we have the following properties:
\begin{itemize}
    \item Suppose we pick $p \in \{p_1, p_2\}$ uniformly at random, and generate a sample set of size $s$ from $p$. Upon receiving the sample set, no algorithm can distinguish whether the sample set is generated according to $p_1$ or $p_2$ with probability more than $(1+\tau)/2$.  
    
    \item For any $s \leq s_2$, there exists a coupling $C$ between $T_1^{s}$ and $T_2^s$ such that
    $$\E[(T_1^s, T_2^s) \sim C]{\Ham{T_1^s}{T_2^s}} \leq \frac{0.4\,\tau}{\privacyeps}\,.$$
\end{itemize}
Let $\MM:[n]^{s} \rightarrow \RR$ be a $\privacyeps$-private algorithm that receives a sample set of size $s$ from an unknown underlying distribution and produces an outcome in a finite set $R$. Then, for any such $\MM$, $s \leq \max(s_1, s_2)$, and any subset of outcomes $R \subseteq \RR$, we have:
$$\Pr[X \sim p_1^{\otimes s}] {\MM(X) \in R} \leq 1.5 \cdot \Pr[X \sim p_2^{\otimes s}]{\MM(X) \in R} + \tau.$$
\end{lemma}

\begin{proof}
We consider two cases based on whether $s \leq s_1$. If $s \leq s_1$, we use the statistical indistinguishability of $p_1$ and $p_2$. The behavior of $\MM$ should be almost identical for sample sets drawn from these two families due to the following standard trick.  Consider an algorithm $\AA$ that runs as follows: 
Upon receiving a sample set $X$, it invokes $\MM(X)$. If the outcome of $\MM$ is in $R$ it declares $p_1$; otherwise, it declares $p_2$. Since this algorithm cannot distinguish $p_1$ from $p_2$, we have: 

\begin{align*}
    \frac{1 + \tau}{2}   \geq \Pr[X\sim p^{\otimes {s}}]{\AA(X) = p} & = \Pr[X\sim p^{\otimes {s}}]{\AA(X) = p_1| p = p_1} \cdot \Pr{p = p_1} 
    \\ & + \Pr[X\sim p^{\otimes {s}}]{\AA(X) = p_2| p = p_2} \cdot \Pr{p = p_2} 
    \\ & = \frac{1}{2}\left(\Pr[X\sim p_1^{\otimes {s}}]{\MM(X) \in R} + \Pr[X\sim p_2^{\otimes {s}}]{\MM(X) \not \in R}\right) 
    \\ & = \frac{1}{2}\left(
    \Pr[X\sim p_1^{\otimes {s}}]{\MM(X) \in R} 
    + 1 - \Pr[X\sim p_2^{\otimes {s}}]{\MM(X)  \in R}
    \right)
\end{align*}
Thus, we obtain: 
$$\Pr[X\sim p_1^{\otimes {s}}]{\MM(X) \in R} \leq \Pr[X\sim p_2^{\otimes {s}}]{\MM(X)  \in R} + \tau$$

Now, consider the case where $s > s_1$. Since $s \leq \max(s_1, s_2)$, we must have $s \leq s_2$. Thus, we use the coupling $C_2$. Let $W_2$ denote the event that $\Ham{T_1^s}{T_2^s} \geq 0.4/\privacyeps$ for $(T_1^{s}, T_2^{s})$ drawn from $C_2$. By assumption and Markov’s inequality, the probability of $W_2$ is at most $\tau$.  

Using the definition of a $\privacyeps$-private algorithm $\MM$, for every pair of datasets $T_1^{s}$ and $T_2^{s}$ with $\Ham{T_1^s}{T_2^s} \leq 0.4/\privacyeps$, we have  
\begin{align*}
    \Pr{\MM(T_1^{s}) \in R}  
    & \leq e^{\privacyeps \cdot \Ham{T_1^s}{T_2^s}} \,\Pr{\MM(T_2^{s}) \in 
    R}
    \\ & \leq e^{0.4} \cdot  \Pr{\MM(T_2^{s}) \in 
    R} \leq 1.5 \cdot \Pr{\MM(T_2^{s}) \in 
    R}.
\end{align*}

Now, we obtain  
\begin{align*}
    \Pr[X \sim p_1^{\otimes {s}}]{\MM(X) \in R} 
    & =\Pr[(T_1^{s}, T_2^{s}) \sim C_1]{\MM(T_1^{s}) \in R} 
    \\ & = \Pr[(T_1^{s}, T_2^{s}) \sim C_1]{\MM(T_1^{s}) \in R \mid \overline{W_2}} \cdot \Pr[(T_1^{s}, T_2^{s}) \sim C_1]{\overline{W_2}}
    \\ & + 
    \Pr[(T_1^{s}, T_2^{s}) \sim C_1]{\MM(T_1^{s}) \in R \mid W_2} \cdot \Pr[(T_1^{s}, T_2^{s}) \sim C_1]{W_2}
    \\ & \leq 1.5 \cdot \Pr[(T_1^{s}, T_2^{s}) \sim C_1]{\MM(T_2^{s}) \in R \mid \overline{W_2}} \cdot \Pr[(T_1^{s}, T_2^{s}) \sim C_1]{\overline{W_2}}
    \\ & + 
    \left( 1.5 \cdot \Pr[(T_1^{s}, T_2^{s}) \sim C_1]{\MM(T_2^{s}) \in R \mid T_1^{s}\not =  T_2^{s}} + 1\right) \cdot  \Pr[(T_1^{s}, T_2^{s}) \sim C_1]{W_2}
    \\ & 
    \leq 
    1.5 \cdot \Pr[(T_1^{s}, T_2^{s}) \sim C_1]{\MM(T_2^{s}) \in R} + \tau
    \\ & = 1.5 \cdot \Pr[X \sim p_2^{\otimes {s}}]{\MM(X) \in R}  + \tau
    \,.
\end{align*}
Hence, the proof is complete.  
\end{proof}

\subsection{Construction of Hard Distributions}

It remains to construct three distributions which satisfy the indistinguishability and closeness assumptions of Lemma~\ref{lem:pairwise_indistinguishability}. We will use a similar construction to \cite{aliakbarpourIRS24}. Some of the necessary properties are well-known; we will prove the rest.

\paragraph{Choice of distributions:} First, we describe the three distributions used in the proof. For the following constructions, we will assume without loss of generality that $n$ is even. (For odd $n$, we can set the probability of the last element to 0 and choose the remaining elements using the construction for $n-1$.) Let $q =  U_n$ be the uniform distribution on $[n]$. We will define the predicted distribution $\hatp$ as follows:
\begin{align}
    \hatp_i \coloneqq \begin{cases}
                    \frac{1+2\eta}{n}, & \text{$i$ is even} \\
                    \frac{1-2\eta}{n}, & \text{$i$ is odd} \\
                \end{cases}
\end{align}
Note that $\totalvardist{\hatp}{q} = \eta$. 

We will generate two other distributions $p^\bullet$ and $p^\diamond$. Set $\testingeps' \in (0,1/2]$ to be $\Theta(\testingeps)$ and $\accuracy' \in (0, \accuracy)$ such that $\eta-\accuracy' = \Theta(\eta - \accuracy)$. Let $\mathbf{Z} \sim \text{Unif}(\{-1,1\}^{n/2})$. Define $p^\bullet$ and $p^\diamond$ as follows:
\begin{align}
    \label{eq:pdot}
    p_i^\bullet = \begin{cases}
                    \frac{1+2\mathbf{Z}_{i/2} \cdot \testingeps'}{n}, & \text{$i$ is even} \\
                    \frac{1-2\mathbf{Z}_{i/2} \cdot \testingeps'}{n}, & \text{$i$ is odd} \\
                \end{cases}
\end{align}
\begin{align}
    \label{eq:pdiamond}
    p_i^\diamond = \begin{cases}
                    \frac{1+2 \cdot (\eta - \accuracy')}{n}, & \text{$i$ is even} \\
                    \frac{1-2 \cdot (\eta - \accuracy')}{n}, & \text{$i$ is odd} \\
                \end{cases}
\end{align}
Note that $p^\bullet$ is $\testingeps'$-far from $U_n$, so $\accept$ is an incorrect answer given samples from $p^\bullet$. Similarly, $p^\diamond$ is $\accuracy$-close to $\hatp$, so $\invalid$ is an invalid answer given samples from $p^\diamond$. Finally, $\reject$ is an invalid answer given samples from $U_n$. Therefore, no single answer works for all three distributions $U_n$, $p^\bullet$, and $p^\diamond$. \\

To prove the privacy assumption of Lemma~\ref{lem:pairwise_indistinguishability}, we will need two couplings: a coupling $\mathcal{C}^\bullet$ between $U_n$ and $p^\bullet$, and a coupling $\mathcal{C}^\diamond$ between $U_n$ and $p^\diamond$. The existence of $\mathcal{C}^\bullet$ follows from a known result:
\begin{lemma}[\cite{AcharyaSZ18}]
    \label{lem:paninskicoupling}
    Let $\testingeps' \in (0,1/2)$, and $n > 0$ be an even number. Let $U_n$ be the uniform distribution on $[n]$, and let $p^\bullet$ be the distribution defined in~\cref{eq:pdot}. Then there exists a coupling $\mathcal{C}^\bullet$ between $U_n^{\otimes s}$ and $p^{\bullet \otimes s}$ such that for any $(T_1^s, T_2^s) \sim \mathcal{C}^\bullet$, we have
    \begin{equation}
        \E{\Ham{T_1^s}{T_2^s}} \leq C(\testingeps')^2 \cdot \min \left( \frac{s^2}{n}, \frac{s^{3/2}}{n^{1/2}} \right)
    \end{equation}
    for some sufficiently large constant $C>0$. 
\end{lemma}

Finally, we will show that there exists a coupling between $T_1^s \sim U_n^{\otimes s}$ and $T_3^s \sim \mathcal{C}^{\diamond \otimes s}$ such that the Hamming distance between the two is small. 
\begin{lemma}
    \label{lem:coincoupling}
    Let $\eta, \accuracy' > 0$ be parameters such that $\eta-\accuracy' \in (0,1/2)$, and $n > 0$ be an even number. Let $U_n$ be the uniform distribution on $[n]$, and let $p^\diamond$ be the distribution defined in ~\cref{eq:pdiamond}. Then there exists a coupling $\mathcal{C}^\diamond$ such that for any $(T_1^s, T_3^s) \sim \mathcal{C}^\diamond$, we have
    \begin{equation}
        \E{\Ham{T_1^s}{T_3^s}} = s(\eta-\accuracy')
    \end{equation}
\end{lemma} 
\begin{proof}
    We will construct the coupling by the following process: 
    \begin{enumerate}
        \item Draw $P_1,...,P_s \sim \text{Unif}([n/2])$.  
        \item For each $j \in [s]$, draw $Z_j \sim \Bern{1/2}$ and $Z_j' \sim \Bern{1 - 2(\eta-\accuracy')}$. Let $X_j = 2P_j - Z_j$. Then define $Y_j$ as 
        \begin{align*}
            Y_j = \begin{cases}
                2P_j - Z_j, & Z_j = 0 \\
                2P_j - Z_j', & Z_j = 1
            \end{cases}
        \end{align*}
        \item Let $T_1^s = \{X_1,...,X_n\}$ and $T_3^s = \{Y_1,...,Y_n\}$. Return $(T_1^s, T_3^s)$.
    \end{enumerate}
    First, we will show that this is a valid coupling: that is, the marginal distributions on $T_1^s$ and $T_3^s$ are $U_n^{\otimes s}$ and $p^{\diamond \otimes s}$ respectively. For all $i \in [n]$ and $j \in [s]$,
    \begin{equation*}
        \Pr{X_j = i} = \frac{1}{2} \cdot \Pr { P_j = \left\lceil \frac{i}{2} \right\rceil} = \frac{1}{n} 
    \end{equation*}
    Since the values $X_j$ are independent, the marginal on $T_1^s$ is $U_n^{\otimes s}$. For all even $i \in [n]$ and $j \in [s]$, we also have
    \begin{align*}
        \Pr{Y_j = i} &= \Pr { P_j = \frac{i}{2}  } \cdot \left( \Pr{Z_j = 0} + \Pr { Z_j' = 0 \mid Z_j = 1 } \cdot \Pr{Z_j = 1} \right) \\
        &= \frac{2}{n} \left( \frac{1}{2} + 2(\eta-\accuracy) \cdot \frac{1}{2} \right) \\
        &= \frac{1+2(\eta-\accuracy')}{n}
    \end{align*}
    For all odd $i$, 
        \begin{align*}
        \Pr{Y_j = i} &= \Pr { P_j = \left\lceil \frac{i}{2} \right\rceil } \cdot \Pr {Z_j' = 1 \mid Z_j = 1 } \cdot \Pr{Z_j = 1} \\
        &= \frac{2}{n} \cdot (1-2(\eta-\accuracy')) \cdot \frac{1}{2} \\
        &= \frac{1-2(\eta-\accuracy')}{n}
    \end{align*}
    which proves that $\mathcal{C}^\diamond$ is a valid coupling. 

    Finally, we bound the expected Hamming distance between the two sample sets. For all $j$, $X_j \neq Y_j$ if and only if $Z_j = 1$ and $Z_j' = 0$. Therefore,
    \begin{align*}
        \E{\Ham{T_1^s}{T_3^s}} &= \E{\sum_{j=1}^s \mathds{1} [X_j \neq Y_j]} 
        = \sum_{j=1}^s \Pr{X_j \neq Y_j} \\
        &= \sum_{j=1}^s \Pr{Z_j = 1} \cdot \Pr{Z_j' = 0} \\
        &= \sum_{j=1}^s \frac{1}{2} \cdot 2(\eta -\accuracy') \\
        &= s(\eta-\accuracy')
    \end{align*}
    as desired.
\end{proof}

\subsection{Proof of~\cref{thm:idlowerbound}}

\begin{proof}
    Suppose $\AA\colon [n]^s \to \{\accept, \reject, \invalid\}$ is an $(\accuracy, \testingeps, \delta=0.2)$-augmented identity tester which uses
    \begin{equation}
        \label{eq:contradiction}
        s < \min \left( \max\left( \frac{0.004\sqrt{n}}{(\testingeps')^2}, \frac{C_1\sqrt{n}}{\testingeps'\sqrt{\privacyeps}}, \frac{C_2n^{1/3}}{(\testingeps')^{4/3}\privacyeps^{2/3}} \right), \max \left( \frac{0.00005}{(\eta-\accuracy')^2}, \frac{0.4}{\xi(\eta-\accuracy')} \right) \right) 
    \end{equation}
    for some constants $C_1,C_2$. We will use Lemma~\ref{lem:pairwise_indistinguishability} and the distributions constructed above to show that $\AA$ must have a high error probability if its sample complexity is too low. Let $T_1^s \sim U_n^{\otimes s}$, $T_2^s \sim p^{\bullet \otimes s}$, and $T_3^s \sim p^{\diamond \otimes s}$. We have the following facts about the distributions:
    \begin{itemize}
        \item By a result of \cite{Paninski08}, no algorithm can distinguish $U_n$ from $p^\bullet$ with probability greater than $0.505$ with fewer than $0.004\sqrt{n}/(\testingeps')^2$ samples.
        
        \item By Lemma~\ref{lem:paninskicoupling}, for any $s \leq \max( C_1\sqrt{n}/(\testingeps'\sqrt{\privacyeps}), C_2n^{1/3}/((\testingeps')^{4/3}\privacyeps^{2/3}) )$ there exists a coupling $\mathcal{C}^\bullet$ between $T_1^s$ and $T_2^s$ such that 
        \begin{equation*}
            \E[(T_1^s,T_2^s) \sim \mathcal{C}^\bullet]{\Ham{T_1^s}{T_2^s}} \leq \frac{0.4}{\xi}
        \end{equation*}
        
        \item No algorithm can distinguish $U_n$ from $p^\bullet$ with probability greater than $0.505$ with fewer than $0.00005/(\eta - \accuracy')^2$ samples. This follows by noting that distinguishing a distribution which is biased toward even elements from the uniform distribution is equivalent to distinguishing a $\eta-\accuracy'$-biased coin from a fair one. It is a folklore fact that this requires at least $\Omega(1/(\eta-\accuracy')^2$ samples. (See, e.g., \cite{aliakbarpourIRS24}). 
        
        \item By Lemma~\ref{lem:coincoupling}, if $s \leq 0.4/(\xi(\eta-\accuracy'))$, there exists a coupling $\mathcal{C}^\diamond$ between $T_1^s$ and $T_3^s$ such that 
        \begin{equation*}
            \E[(T_1^s,T_3^s) \sim \mathcal{C}^\diamond]{\Ham{T_1^s}{T_3^s}} \leq \frac{0.4}{\xi}
        \end{equation*}
    \end{itemize}
    Since $\AA$ has only three possible outputs, we have: 
    \begin{align*}
        1 & = \Pr[T_1 \sim U_n^{\otimes {s}}]{\AA(T_1) = \accept}
         + \Pr[T_1 \sim U_n^{\otimes {s}}]{\AA(T_1) = \reject}
         +\Pr[T_1 \sim U_n^{\otimes {s}}]{\AA(T_1) = \invalid}
    \end{align*}
    Using Lemma~\ref{lem:pairwise_indistinguishability}, we have:
    \begin{align*}
        1 & \leq \Pr[T_1 \sim U_n^{\otimes {s}}]{\AA(T_1) = \accept}
         + 1.5 \cdot \Pr[T_2 \sim p^{\dot \otimes {s}}]{\AA(T_2) = \reject} 
         + 1.5 \cdot \Pr[T_3\sim p^{\diamond \otimes {s}}]{\AA(T_3) = \invalid} + 0.02
    \end{align*}
    Based on our definition of $U_n$,  $p^\bullet$ and $p^\diamond$, the three probabilities above are at most the probability that $\AA$ makes a mistake. Thus, we have: 
    $$1 \leq 4 \delta + 0.02\,$$
    Hence, $\delta \geq 0.245$, which contradicts our assumption that $\delta = 0.2$. Therefore, $s$ cannot be lower than the bound in~\cref{eq:contradiction}, which immediately implies the lower bound in~\cref{eq:idlowerbound}. 
\end{proof}

\section{Upper bound for Private Augmented Closeness Testing} 
\label{sec:UB_closeness}

Our result for closeness testing combines the flattening-based closeness testers of \cite{aliakbarpourIRS24} and \cite{ADKR19}. The flattening technique, introduced by \cite{DiakonikolasK16}, uses samples to map the original distributions $p$ and $q$ onto new distributions $p'$ and $q'$ such that $p'$ has a low $\ell_2$ norm, but $\totalvardist{p}{q} = \totalvardist{q'}{q'}$. The main result of \cite{ADKR19} is that any flattening procedure which satisfies certain technical conditions can be made private. Since the augmented closeness tester of \cite{aliakbarpourIRS24} also relies on flattening, we show that it too can be made private with some modifications. In~\cref{sec:flatteningov}, we review the flattening procedures of \cite{aliakbarpourIRS24} and \cite{ADKR19}. We describe our own procedure in~\cref{sec:flattening}, and present our algorithm in~\cref{sec:closenessalg}. Finally, in~\cref{sec:l2estimation}, we show how to privately estimate the $\ell_2$ norm to verify the success of our flattening. 

\subsection{Overview of Flattening}
\label{sec:flatteningov}

Flattening is motivated by a result of \cite{ChanDVV14}, in which the authors give a closeness tester for distributions $p$ and $q$ whose sample complexity is $O(bn/\testingeps^2)$, where $b = \min(\|p\|_2, \|q\|_2)$. The goal is to map $p$ and $q$ to distributions with lower $\ell_2$ norm while preserving their distance: as a guiding example, if one could map either $p$ or $q$ to (something close to) the uniform distribution on $\Theta(n)$ elements, then one could plug $b = O(1/\sqrt{n})$ in the above sample complexity. 

The flattening technique described by \cite{DiakonikolasK16} works as follows: draw a multiset $F$ of $\Poi{k}$ samples from $p$, and, for each $i\in[n]$, denote by $k_i$ the number of elements in $F$ equal to $i$. The new distribution $p'$ will be supported on domain $\{(i,j) : i \in [n], j \in [k_i+1] \}$. To draw a sample from $p'$, draw $i$ from $p$, choose $j$ uniformly at random from the set $[k_i+1]$, and output $(i,j)$. This divides the probability mass of each element $i$ evenly into $k_i+1$ buckets. Diakonikolas and Kane \citep{DiakonikolasK16} show that the expected $\ell_2$ norm of $p'$ is low, namely,
\begin{equation*}
    \mathbf{E}\left[ \left\|p'\right\|_2^2 \right] \leq \frac{1}{k}
\end{equation*}
On the other hand, one can easily check that $\totalvardist{p'}{q'} = \totalvardist{p}{q}$, and that the new domain has size $n' = O(n+k)$. Therefore, choosing $k$ carefully, we can now draw a new set of test samples and test the closeness of $p'$ and $q'$ at a much lower cost than testing that of $p$ and $q$ directly. 

In the rest of this section, we describe two variations on the original technique which will be useful in constructing our augmented private tester.

\paragraph{Private Flattening:} The private tester of \cite{ADKR19} attempts to derandomize the original flattening of \cite{DiakonikolasK16}. When a flattening-based tester divides its samples into flattening and test sets, it introduces randomness based on the permutation $\pi$ used to partition the samples. This results in high sensitivity of the test statistic $Z$, as swapping one of the samples used for the flattening can wildly affect the resulting value of $Z$. The private tester reduces the sensitivity by taking the expectation of $Z$ over every permutation $\pi$ of the data and the randomness $r$ of the flattening procedure:
\begin{equation*}
    \overline{Z} \coloneqq \E[\pi,r]{Z}
\end{equation*}
The authors show that $\overline{Z}$ has low sensitivity and preserves the correctness of any flattening procedure which satisfies two technical conditions. Such a flattening is called a \textit{proper procedure}. 
\begin{definition}
    \label{def:proper}
    Let $\AA$ be an algorithm which draws a multiset $X$ from distribution $p$ and reduces testing property $\PP$ of $p$ to testing closeness between two distributions $p'$ and $q'$. We say $\AA$ is a \textit{proper procedure} if there exist  constants $c_0 < 1$ and $c_1 \geq 1$ such that the following holds for any $p'$ and $q'$ produced by $\AA$:
    \begin{equation}
        \label{eqn:permutation bound}
        \Pr[X]{\mathbf{E}_\pi\!\left[ \left\| p' - q' \right\|_2^2 \,\bigg\vert\, X \right] \geq 4 c_0 \cdot \mathbf{E}_{\AA}\!\left[ \left\| p' - q' \right\|_2^2  \right]  } \geq 9/10 
    \end{equation}
    \begin{equation}
        \label{eqn:fourthmomentbound}
        \mathbf{E}_\AA \!\left[ \left\| p' - q' \right\|_4^4  \right] \leq c_1  \cdot \left( \mathbf{E}_\AA \!\left[ \left\| p' - q' \right\|_2^2  \right] \right)^2
    \end{equation}
\end{definition}
As we show in~\cref{sec:flattening}, a crucial step in our flattening procedure is identical to one proven to be proper in \cite{ADKR19}: the flattening of \cite{DiakonikolasK16} when the flattening set is drawn from both $p$ and $q$. 
\begin{lemma}
    \label{lem:properprocedure}
    Let $\AA$ be an algorithm for testing closeness of distributions $p$ and $q$ with the following flattening procedure: draw a multiset $F^{(p)}$ from $p$ and a multiset $F^{(q)}$ from $q$. Let $k_i$ denote the number of instances of element $i$ in $F^{(p)} \cup F^{(q)}$. Divide the probability mass of element $i$ evenly into $k_i$ buckets. Then $\AA$ is a proper procedure.
\end{lemma}
The sample complexity of any proper procedure is given by the following result. 
\begin{lemma}[\cite{ADKR19}]
    \label{lem:permutationtester}
    Let $\mathcal{A}$ be a proper procedure for testing property $\mathcal{P}$ which reduces the problem to testing closeness of $p'$ and $q'$ over a domain of size $n'$ using $\Poi{k}$ flattening samples. Then there exists an $(\privacyeps, \testingeps, 3/4)$-private tester for property $\mathcal{P}$ which draws $\Poi{s}$ test samples, for any
    \begin{equation}
        s \geq \Theta \left( \frac{n' \cdot \sqrt{\mathbf{E}_\AA \left[ \min(\|p'\|_2^2, \|q'\|_2^2) \right]}}{\testingeps^2} + \frac{\sqrt{n\Delta(\overline{Z})}}{\testingeps\sqrt{\privacyeps}} \right)
    \end{equation}
    Additionally, the sensitivity of the test statistic $\overline{Z}$ is bounded by
    \begin{equation}
        \Delta(\overline{Z}) = O \left( \frac{s+k}{k} \right) 
    \end{equation}
\end{lemma}

\paragraph{Augmented Flattening:} In the augmented setting of \cite{aliakbarpourIRS24}, the estimate $\hat{p}$ guides the choice of buckets for each element $i$. Let $\nu \in (0, 1]$ be a flattening parameter. Rather than dividing elements into $k_i+1$ buckets, the flattener creates
\begin{equation}
    \label{eqn:augmentedflattening}
    b_i = \left\lfloor \frac{\hat{p}_i}{\nu} \right\rfloor + k_i + 1
\end{equation}
buckets in flattened distributions $p'$ and $q'$. If $\hat{p}$ is $\accuracy$-close to $p$, the expected $\ell_2$ norm after flattening is shown to be
\begin{equation}
    \mathbf{E} \left[ \left\|p'\right\|_2^2 \right] \leq \frac{2\accuracy}{k} + 4 \cdot \nu
\end{equation}
The tester then draws $O(\sqrt{n})$ samples each from $p'$ and $q'$ to estimate their $\ell_2^2$ norm. If at least one distribution has been sufficiently flattened, the algorithm proceeds to the test phase. Otherwise, it returns $\invalid$. The sample complexity of this tester is given below. 

\begin{lemma}[\cite{aliakbarpourIRS24}]
    Fix parameters $\alpha, \testingeps \in (0,1]$. Let $p$ and $q$ be unknown distributions, and $\hat{p}$ be a known distribution over $[n]$. There exists an $(\testingeps, \alpha, \errprob=0.3)$-augmented closeness tester for $p$ and $q$ which uses
    $
        \Theta \left( \frac{n^{2/3}\accuracy^{1/3}}{\testingeps^{4/3}} + \frac{\sqrt{n}}{\testingeps^2} \right)
    $
    samples from both $p$ and $q$. 
\end{lemma}
As discussed in the introduction, to mimic this approach we need to perform this last step~--~the estimation of the $\ell_2^2$ norm to verify that the flattening was successful~--~in a \emph{differentially private} fashion.

\subsection{Two-Step Flattening}
\label{sec:flattening}

To simplify the analysis of our augmented private closeness tester, we split the flattening into two steps. The first step flattens $p$ and $q$ into distributions $p'$ and $q'$ using only the prediction $\hat{p}$. The second step draws samples from $p'$ and $q'$ and uses these samples to flatten them into a third set of distributions $p''$ and $q''$. This two-step flattening makes it much easier to guarantee that the procedure is private. Since the first step draws no samples, it is, as-is, already differentially private. Since the second step does not use $\hat{p}$, it is precisely the same procedure as the flattening described in Lemma~\ref{lem:properprocedure}, and can be made private by the same mechanism. We show that if $\totalvardist{\hat{p}}{p} \leq \accuracy$, then $\|p''\|_2^2$ is small in expectation, allowing us to test efficiently with fewer flattening samples than the optimal non-augmented tester. 

\paragraph{Flattening procedure:} Let $k$ be a parameter which will determine the expected number of flattening samples. The two-step flattening works as follows. 
\begin{enumerate}
    \item From the original distributions $p$ and $q$, estimate $\hat{p}$, and suggested accuracy $\accuracy$, construct $p'$ and $q'$ by flattening each element $i\in[n]$ into $b_i$ buckets, where
    \begin{equation}
        \label{eqn:flatten1}
        b_i = \lceil n\hat{p}_i \rceil + 1
    \end{equation}
    (This is analogous to setting $\nu = 1/n$ in Equation \ref{eqn:augmentedflattening}). Since $\hat{p}$ and $\accuracy$ are not private, this can be done as a preprocessing step before running the private portion of the algorithm. After this step, $p'$ is defined by having, for all $i \in [n]$ and $j \in [b_i]$, $p_{i,j}' = p_i/b_i$. Sampling from $p'$ can be done as follows: sample $i \sim p$ and $j \sim \text{Unif}([b_i])$, and return $(i,j)$. Sampling from $q'$ is analogous (note that both $p$ and $q$ use the same bucketing, so $p'$ and $q'$ have the same domain).

    \item Sample $\hat{k}_p,\hat{k}_q \sim \Poi{k}$ independently. Draw a multiset $F^{(p)}$ of $\hat{k}_p$ samples from $p'$ and a multiset $F^{(q)}$ of $\hat{k}_q$ samples from $q'$. Let $F = F^{(p)} \cup F^{(p)}$, and $k_{i,j}$ be the total number of instances of element $(i,j)$ in $F$. Then divide each element of the domain of $p',q'$ into $b_{i,j} = k_{i,j}+1$ buckets. 

    The domain of $p''$ and $q''$ is $\Omega = \{(i,j,m): i \in [n], j \in [b_i], m \in [b_{i,j}]\}$, and 
    \begin{equation}
        \label{eqn:flatten2}
    p_{i,j,m}'' = \frac{p_{i,j}}{b_{i,j}}, \qquad q_{i,j,m}'' = \frac{q_{i,j}}{b_{i,j}}
    \end{equation}
    for all $(i,j,m)\in \Omega$. To sample from $p''$, we sample $(i,j) \sim p'$, $m \sim \text{Unif}([b_{i,j}])$, then return $(i,j,m)$. 

\end{enumerate}

\paragraph{Effect of flattening:} We now show that the domain size after flattening remains $\Theta(n+k)$ with high probability. After the second flattening step, the number of elements is $n'' = |\Omega|$ (itself a random variable), where
\begin{equation*}
    n'' = \sum_{i=1}^n \sum_{j=1}^{b_i} b_{i,j} = \sum_{i=1}^n \sum_{j=1}^{b_i} (k_{i,j} + 1) = \hat{k}_p + \hat{k}_q + \sum_{i=1}^n b_i \leq \hat{k}_p + \hat{k}_q + \sum_{i=1}^n (n\hat{p}_i + 2) \leq \hat{k}_p + \hat{k}_q + 3n
\end{equation*}
Note that $\hat{k}_p + \hat{k}_q \sim \Poi{2k}$, so $\hat{k}_p + \hat{k}_q \leq 20k$ with probability at least $9/10$ by Markov's inequality. Therefore, $n'' = O(n+k)$ with probability at least $9/10$. Next, we bound the expected $\ell_2^2$ norm of the flattened distribution, assuming the suggested accuracy level $\accuracy$ is valid.

\begin{lemma}[{See, e.g., \cite[Lemma 2.6]{DiakonikolasK16}}]
    \label{lemma:poisson:expectation:inverse}
    Fix $\lambda > 0$. If $X\sim \Poi{\lambda}$, then $\E{\frac{1}{1+X}} \leq \frac{1}{\lambda}$.
\end{lemma}
\begin{proof}
    Follows from manipulating the series corresponding to the expectation:
    \[
    \E{\frac{1}{1+X}} = e^{-\lambda}\sum_{k=0}^\infty \frac{1}{k+1}\frac{\lambda^k}{k!}
    = \frac{1}{\lambda}e^{-\lambda}\sum_{k=0}^\infty \frac{\lambda^{k+1}}{(k+1)!}
     = \frac{1}{\lambda}e^{-\lambda}\sum_{k=1}^\infty \frac{\lambda^{k}}{k!}
    = \frac{1-e^{-\lambda}}{\lambda}\,. \qedhere
    \]
\end{proof}

\begin{lemma}
    \label{lemma:twostepflattening}
    Let $p$ and $\hat{p}$ be distributions over $[n]$ with $\totalvardist{p }{\hat{p}} \leq \accuracy$. Then the two-step flattening produces a distribution $p''$ with 
    \begin{equation}
        \mathbf{E}_F \left[ \|p''\|_2^2 \right] \leq \frac{2\accuracy}{k} + \frac{4}{n}
    \end{equation}
\end{lemma}
\begin{proof}
    The proof is inspired by \cite{aliakbarpourIRS24}. Let $\Delta_{i} = p_{i} - \hat{p}_{i}$ for all $i\in[n]$, so that 
    \begin{equation*}
        p_i = \Delta_i + \hat{p}_i \leq 2\max(\Delta_i,\hat{p}_i)
    \end{equation*}
    Letting $A \coloneqq \{i \in [n] : \Delta_i \geq \hat{p}_i \}$, this gives the following bound on $p_i$:
    \begin{equation*}
        p_i \leq 
        \begin{cases}
            2\Delta_i, &\text{ } i \in A \\
            2\hat{p}_i, &\text{ } i \in [n] \setminus A \\
        \end{cases}
    \end{equation*}
    By definition of $p'$ and $p''$, we have
    \begin{align*}
        \|p''\|_2^2 
        &= \sum_{i=1}^n \sum_{j=1}^{b_i} \sum_{m=1}^{b_{i,j}} (p_{i,j,m}'')^2 = \sum_{i=1}^n \sum_{j=1}^{b_i} \frac{(p_{i,j}')^2}{b_{i,j}} \\
        &= \sum_{i \in A} \sum_{j=1}^{b_i} \frac{(p_{i,j}')^2}{b_{i,j}} + \sum_{i \in [n] \setminus A} \sum_{j=1}^{b_i} \frac{(p_{i,j}')^2}{b_{i,j}} \\
        &\leq \sum_{i \in A} \sum_{j=1}^{b_i} \frac{2\Delta_ip_{i,j}'}{b_ib_{i,j}} + \sum_{i \in [n] \setminus A} \sum_{j=1}^{b_i} \frac{p_i^2}{b_i^2 b_{i,j}} \\
        &\leq \sum_{i \in A} \sum_{j=1}^{b_i} \frac{2\Delta_ip_{i,j}'}{b_ib_{i,j}} + \sum_{i \in [n] \setminus A} \sum_{j=1}^{b_i} \frac{p_i^2}{b_i^2} \tag{$b_{i,j} \geq 1$}\\
        &\leq \sum_{i \in A} \sum_{j=1}^{b_i} \frac{2\Delta_ip_{i,j}'}{b_i(k_{i,j}+1)} + \sum_{i \in [n] \setminus A} \sum_{j=1}^{b_i} \frac{(2\hat{p}_i)^2}{b_i^2} \\
         &\leq \sum_{i \in A} \sum_{j=1}^{b_i} \frac{2\Delta_ip_{i,j}'}{b_i(k_{i,j}+1)} + 4\sum_{i \in [n] \setminus A}  \frac{\hat{p}_i^2}{b_i} \\
         &\leq \sum_{i \in A} \sum_{j=1}^{b_i} \frac{2\Delta_ip_{i,j}'}{b_i(k_{i,j}+1)} + 4\sum_{i \in [n] \setminus A}  \frac{\hat{p}_i^2}{n\hat{p}_i} \tag{$b_i \geq n \hat{p}_i$} \\
         &\leq \sum_{i \in A} \sum_{j=1}^{b_i} \frac{2\Delta_ip_{i,j}'}{b_i(k_{i,j}+1)} + \frac{4}{n} 
    \end{align*}
    Note that we can write $k_{i,j} = k^{(p)}_{i,j} + k^{(q)}_{i,j}$, where $k^{(p)}_{i,j} \sim \Poi{kp'_{i,j}}$ and $k^{(q)}_{i,j} \sim \Poi{kq'_{i,j}}$. Applying Lemma~\ref{lemma:poisson:expectation:inverse},
    \begin{equation*}
        \mathbf{E}_F \left[ \frac{1}{k_{i,j}+1} \right] \leq \mathbf{E}_F \left[ \frac{1}{k^{(p)}_{i,j}+1} \right] \leq \frac{1}{kp'_{i,j}}
    \end{equation*}
    Combining the two results, the expected norm is bounded by
    \begin{equation*}
        \mathbf{E}_F \left[ \|p''\|_2^2 \right] \leq \sum_{i \in A} \left( \frac{2\Delta_i}{k} \right) + \frac{4}{n} \leq \frac{2\accuracy}{k} + \frac{4}{n}
    \end{equation*}
    the last inequality since $\sum_{i\in A} \Delta_i = p(A) - \hat{p}(A) \leq \sup_{S}(p(S)-\hat{p}(S)) = \totalvardist{p}{\hat{p}} \leq \accuracy$.
\end{proof}

\subsection{The Algorithm}
\label{sec:closenessalg}
At a high level, our algorithm works in three steps. The first step flattens $p$ and $q$ into distributions $p''$ and $q''$ as described in~\cref{sec:flattening}. The second step tests the $\ell_2^2$ norm of $p''$. If the norm is not within a constant multiplicative factor of the bound guaranteed by Lemma~\ref{lemma:twostepflattening}, then we have $\totalvardist{\hat{p}}{p} > \accuracy$ with high probability, and the tester returns $\invalid$. If the norm \textit{is} close to the bound, our algorithm can perform an efficient closeness test. The final step runs the private closeness tester of \cite{ADKR19} and returns its result. We give pseudocode for our algorithm in Algorithm \ref{alg:closenesstester}. 

By our lower bound (\cref{thm:closenesslowerbd}), there are some regimes of parameters in which an augmented tester cannot use asymptotically fewer samples than a non-augmented private tester. In these cases, we simply invoke the non-augmented tester, whose sample complexity is given by Lemma~\ref{lem:privateclosenesstesting}. In all other regimes, we show that our tester is optimal up to $\log n$ factors.

%
%
\begin{lemma}[\cite{ZhangThesis21}]
    \label{lem:privateclosenesstesting}
    Let $\testingeps \in (0,1]$ and $\privacyeps > 0$. Let $p$ and $q$ both be unknown distributions over $[n]$. Then there is a  $(\privacyeps, \testingeps, 0.05)$-private closeness tester for $p$ and $q$ which takes 
    $
        \Theta \left( \frac{n^{2/3}}{\testingeps^{4/3}} + \frac{\sqrt{n}}{\testingeps^2} + \frac{1}{\testingeps\privacyeps} + \frac{\sqrt{n}}{\testingeps\sqrt{\privacyeps}} + \frac{n^{1/3}}{\testingeps^{4/3}\privacyeps^{2/3}} \right)
    $
    samples. 
\end{lemma}

We are now ready to prove our upper bound. 
\ifcolt
\else
    \begin{algorithm}[htbp]
\small
\caption{\textsc{Private-Augmented-Closeness-Tester}}\label{alg:closenesstester}
\KwIn{$p, q, \hat{p}, \accuracy, n, \errprob=0.32$}
\KwOut{$\accept$, $\reject$, or $\invalid$} 

\DontPrintSemicolon
\nl\uIf{$\testingeps = o(n^{-1/4})$ or $\testingeps^2\privacyeps = o(n^{-1})$} {
\nl Run the tester of Lemma~\ref{lem:privateclosenesstesting} and return its answer \;
}
\nl $\hat{k}_p, \hat{k}_q \gets \Poi{k}$, $\hat{\ell} \gets \Poi{\ell}$ where $k,\ell$ are determined by
\ifcolt
~\cref{thm:closenessupperbd}\; 
\else
~\cref{thm:closenessupperbd:restated}
\fi

\nl \uIf{$\hat{k} + \hat{\ell} > 10(k+\ell)$} {
    \nl \Return $\reject$ \;
}
\nl $p', q' \gets$ Flatten $p$ and $q$ according to~\cref{eqn:flatten1} \;

\nl $p'', q'' \gets$ Flatten $p'$ and $q'$ according to~\cref{eqn:flatten2} using $\hat{k}_p$ samples from $p$ and $\hat{k}_q$ from $q$\;

\nl $\overline{L} \gets$ compute the statistic in~\cref{eqn:derandomizedl2tester} using $\hat{\ell}$ samples \;

\nl $\tilde{L} \gets \overline{L} + \Lap{\frac{\Delta(\overline{L})}{\privacyeps}}$ \;

\nl \uIf{$\tilde{L} > 30 \cdot \left( \frac{2\accuracy}{k} + \frac{4}{n} \right)$} {
    \nl \Return{$\invalid$} \;
}
\nl Run the tester of Lemma~\ref{lem:permutationtester} and return its answer \label{line:permutationtester} \;
\end{algorithm}
\fi

\begin{theorem}
\ifcolt
[\cref{thm:closenessupperbd}, restated]
\fi
\label{thm:closenessupperbd:restated}
Let $\testingeps, \accuracy \in (0,1]$ and $\xi > 0$. Let $p$ and $q$ both be unknown distributions over $[n]$. Then Algorithm~\ref{alg:closenesstester} is a $\privacyeps$-differentially private augmented $(\testingeps, \accuracy, 0.32)$-closeness tester which takes $s$ samples each from $p$ and $q$, where 
\begin{equation}
    \label{eq:closenesssamplecomplexity}
    s =  \tilde{O} \left( \frac{n^{2/3}\accuracy^{1/3}}{\testingeps^{4/3}} + \frac{\sqrt{n}}{\testingeps^2} + \frac{1}{\testingeps\privacyeps} + \frac{\sqrt{n}}{\testingeps\sqrt{\privacyeps}} + \frac{n^{1/3}}{\testingeps^{4/3}\privacyeps^{2/3}} \right)
\end{equation}
\end{theorem}
\begin{proof}    
    For the analysis, we will condition on two assumptions: 
    \begin{enumerate}
        \item $\hat{k} =  \hat{k}_p+ \hat{k}_q < 100k$, $\hat{\ell} < 100\ell$, and $\hat{s} < 100s$. This holds with probability at least $0.97$ by Markov's inequality and a union bound. 
        \item For all $i \in [n]$,
        \begin{equation}
            \label{eq:sensitivityassumption}
                \frac{\ell_i}{k_i+1} < 12 
                \log \left( \frac{n}{0.05} \right) \cdot \frac{\ell}{k}
        \end{equation}   
        This holds with probability at least $0.95$ by Lemma~\ref{lem:composition}. If this assumption does not hold in our original dataset, we enforce it via the mapping of Lemma~\ref{lemma:randomizedmapping}, which preserves privacy.  
    \end{enumerate}

    \paragraph{Privacy:} If $\testingeps = o(n^{-1/4})$ or $\testingeps^2\privacyeps = o(n^{-1})$, we run the tester of Lemma~\ref{lem:privateclosenesstesting}, which is private by the same result. 
    
    If $\testingeps = \Omega(n^{-1/4})$ and $\testingeps^2\privacyeps = \Omega(n^{-1})$, the algorithm's outputs depend only on the statistic $\tilde{L}$, and the choice of $\hat{k}_p$, $\hat{k}_q$ and $\hat{\ell}$. The latter two do not depend on the sampled data. If assumption (2) is satisfied, then $\tilde{L}$ is $\privacyeps$-
    differentially private by Lemma
    \ifcolt
    ~\ref{lemma:privatel2tester}
    \else 
    ~\ref{lemma:privatel2tester:restated}
    \fi. Finally, the second flattening step is a proper procedure by Lemma~\ref{lem:properprocedure}. Therefore, if the algorithm reaches~\cref{line:permutationtester}, its output is private by Lemma~\ref{lem:permutationtester}.  
    
    \paragraph{Correctness:} If $\testingeps = o(n^{-1/4})$ or $\testingeps^2\privacyeps = o(n^{-1})$, the algorithm returns a valid answer with probability at least $0.95$ by Lemma~\ref{lem:privateclosenesstesting}. Therefore, we need only consider the case where $\testingeps = \Omega(n^{-1/4})$ and $\testingeps^2\privacyeps = \Omega(n^{-1})$. 
    
    First, we show that if $\totalvardist{p}{\hat{p}} \leq \accuracy$, the algorithm does not output $\invalid$ with high probability. The only case in which the algorithm returns $\invalid$ is when $\tilde{L} > 30 (2\accuracy/k + 4/n)$. By Lemma 
    \ifcolt
    \ref{lemma:privatel2tester}
    \else
    \ref{lemma:privatel2tester:restated}
    \fi
    , the estimate $\tilde{L}$ is within a constant factor of $\|p''\|_2^2$ with probability at least 0.94: 
    \begin{equation}
        \frac{\|p''\|_2^2}{2} \leq \tilde{L} \leq 3\frac{\|p''\|_2^2}{2}
    \end{equation}
    Combining Lemma~\ref{lemma:twostepflattening} with Markov's inequality, the following holds with probability at least 0.95:
    \begin{equation}
      \|p''\|_2^2  \leq 20 \left( \frac{2\accuracy}{k} + \frac{4}{n} \right)
    \end{equation}
    Therefore, by union bound, we have that $\tilde{L}$ is not more than 20 times the expected value of $\|p''\|_2^2$ with probability at least $0.91$, and the probability of returning $\invalid$ in this case is at most $0.09$. 

    Next, we show that the algorithm does not output $\reject$ with high probability when $p=q$. Conditioned on assumption (1), the algorithm only rejects if the tester of Lemma~\ref{lem:permutationtester} rejects. This occurs with probability at most $0.25$ when $p = q$. 

    Finally, we show that the algorithm does not return $\accept$ when $\totalvardist{p}{q} > \testingeps$. In this case, the algorithm only accepts if the tester of Lemma~\ref{lem:permutationtester} accepts, which happens with probability at most $0.25$. 

    Therefore, the algorithm outputs a valid answer with probability $0.75$ conditioned on the assumptions. Taking a union bound, the algorithm outputs a valid answer with probability $0.68$ overall. 
    
    \paragraph{Sample Complexity:} By assumption (1), we draw $O(k)$ flattening samples, $O(\ell)$ estimation samples, and $O(s)$ test samples from both $p$ and $q$. Let $n'' = O(n+k)$ be the domain size of $p''$. Then the parameters must satisfy the following constraints:
    \begin{align}
            k \cdot \min \left( \frac{k}{\log^2n}, \frac{\ell}{\log n} \right) = \Omega \left( \frac{n''}{\privacyeps} \right) 
            &\quad \text{(for $\ell_2$ norm estimation)} 
            \label{eq:kconstraint}\\
            \ell \geq \Theta(\sqrt{n''}) 
            &\quad \text{(for $\ell_2$ norm estimation)} 
            \label{eq:lconstraint}\\
            s \geq \Theta \left( \frac{n''\cdot \sqrt{\mathbf{E}_F \left[ \|p''\|_2^2 \right]}}{\testingeps^2} + \frac{\sqrt{n''\Delta(\overline{Z})}}{\testingeps\sqrt{\privacyeps}} \right)  
            &\quad \text{(for testing)}
            \label{eq:sconstraint} 
    \end{align}
    By Lemma~\ref{lem:permutationtester}, the sensitivity of the test statistic is
    \begin{equation*}
        \Delta(\overline{Z}) = \Theta \left( \frac{s+k}{k} \right)\,.
    \end{equation*}
    And by Lemma~\ref{lemma:twostepflattening}, we have
    \begin{equation*}
         \mathbf{E}_F \left[ \|p''\|_2^2 \right] \leq \sqrt{\frac{2\accuracy}{k} + \frac{4}{n}}\,.
    \end{equation*}
    Finally, to simplify the constraint in~\cref{eq:kconstraint}, we note that the equation implies two constraints on $k$:
    \begin{align*}
        \frac{k^2}{\log^2 n} = \Omega \left( \frac{k}{\privacyeps} \right), & \text{ so } k = \Omega \left( \frac{\log^2 n}{\privacyeps} \right) \\
        \frac{k^2}{\log^2 n} = \Omega \left( \frac{n}{\privacyeps} \right), & \text { so } k = \Omega \left( \frac{\sqrt{n}\log n}{\sqrt{\privacyeps}} \right) \,.\\
    \end{align*}
    If both of the above hold and $\ell = k$, the parameters always satisfy~\cref{eq:kconstraint}. Note that the first constraint dominates if and only if $\privacyeps = o(\log n/\sqrt{n})$. 

    We are now ready to describe the sample complexity. We will consider the following cases:
    \begin{description}
        \item[Case 1:] $\testingeps = o(n^{-1/4})$. In this case we run the non-augmented private tester, whose sample complexity is given by Lemma~\ref{lem:privateclosenesstesting}. We have 
        \begin{equation*}
            \frac{n^{2/3}}{\testingeps^{4/3}} = \frac{n^{2/3}\testingeps^{2/3}}{\testingeps^{2}} = O\left(\frac{\sqrt{n}}{\testingeps^{2}}\right)
        \end{equation*}
        Therefore, $n^{2/3}/\testingeps^{4/3}$ is not the dominating term, and the sample complexity can be written as 
        \begin{equation*}
            O \left( \frac{\sqrt{n}}{\testingeps^2} + \frac{n^{1/3}}{\testingeps^{4/3}\privacyeps^{2/3}} + \frac{\sqrt{n}}{\testingeps\sqrt{\privacyeps}} + \frac{1}{\testingeps\privacyeps} \right) 
        \end{equation*}
        which matches our lower bound. 

        \item[Case 2:] $\testingeps = \Omega(n^{-1/4})$ and $\testingeps^2\privacyeps = o(n^{-1})$. Once again, we run the non-augmented tester. By our assumption,
        \begin{equation*}
            \testingeps^2\privacyeps = o(n^{-1}) = o \left( \frac{n^{-2/3}}{n^{1/3}} \right) = o \left( \frac{\testingeps^{8/3}}{n^{1/3}} \right) 
        \end{equation*}
        Rearranging terms, we have $n/\testingeps^2 = o(1/\privacyeps^3)$. Consequently,
        \begin{equation*}
            \frac{n^{2/3}}{\testingeps^{4/3}} = \left( \frac{n}{\testingeps^2} \cdot \frac{n^3}{\testingeps^6} \right)^{1/6} = O \left( \left( \frac{1}{\privacyeps^3} \cdot \frac{n^3}{\testingeps^6} \right)^{1/6} \right) = O \left( \frac{\sqrt{n}}{\testingeps\sqrt{\privacyeps}} \right) 
        \end{equation*}
        Therefore, $n^{2/3}/\testingeps^{4/3}$ is not the dominating term. The sample complexity is  
        \begin{equation*}
            O \left( \frac{\sqrt{n}}{\testingeps^2} + \frac{n^{1/3}}{\testingeps^{4/3}\privacyeps^{2/3}} + \frac{\sqrt{n}}{\testingeps\sqrt{\privacyeps}} + \frac{1}{\testingeps\privacyeps} \right) 
        \end{equation*}
        which matches our lower bound. 

        \item[Case 3:] $\testingeps = \Omega(n^{-1/4})$, $\testingeps^2\privacyeps = \Omega(n^{-1})$, and  
        \begin{equation}
            \label{eq:bestcasesc}
            \frac{\sqrt{n}\log n}{\sqrt{\privacyeps}} = O \left( \frac{n^{2/3}\accuracy^{1/3}}{\testingeps^{4/3}} + \frac{\sqrt{n}}{\testingeps^2} + \frac{\sqrt{n}}{\testingeps\sqrt{\privacyeps}} \right) 
        \end{equation}
        In this case, we run the augmented tester. We set
        \begin{equation}
            \label{eq:augmentedclosenesscomplexity}
            \ell = k = \Theta \left( \frac{n^{2/3}\accuracy^{1/3}}{\testingeps^{4/3}} + \frac{\sqrt{n}}{\testingeps^2} + \frac{\sqrt{n}}{\testingeps\sqrt{\privacyeps}} \right) = O(n)
        \end{equation}
        and
        \begin{equation*}
            s = \Theta \left( \frac{n}{\testingeps^2} \sqrt{\frac{2\accuracy}{k} + \frac{4}{n}}  + \frac{\sqrt{n}}{\testingeps\sqrt{\privacyeps}} \right) = O \left( \frac{n^{2/3}\accuracy^{1/3}}{\testingeps^{4/3}} + \frac{\sqrt{n}}{\testingeps^2} + \frac{\sqrt{n}}{\testingeps\sqrt{\privacyeps}} \right) = O(k)
        \end{equation*}
        Therefore, $\Delta(\overline{Z}) = \Theta(1)$, and $s$ satisfies~\cref{eq:sconstraint}. By our assumptions, we have the following facts:
        \begin{align*}
            \frac{n^{2/3}\accuracy^{1/3}}{\testingeps^{4/3}} &\leq n\accuracy^{1/3} = O(n) \\
            \frac{\sqrt{n}}{\testingeps^2} &= O(n) \\
            \frac{\sqrt{n}}{\testingeps\sqrt{\privacyeps}} &= O \left( n \right)
        \end{align*}
        We also have $\ell = k = \Omega(\sqrt{n}/\testingeps^2) = \Omega(\sqrt{n})$, which satisfies~\cref{eq:lconstraint}. Finally, $k = \Omega(\sqrt{n}\log n / \sqrt{\privacyeps}) = \Omega(\sqrt{n+k}\log n / \sqrt{\privacyeps})$, which satisfies~\cref{eq:kconstraint}. In this case, the overall sample complexity is equal to~\cref{eq:augmentedclosenesscomplexity}. Each of the terms in this upper bound appear in our lower bound, so the sample complexity is optimal. 

        \item[Case 4:] $\testingeps = \Omega(n^{-1/4})$, $\testingeps^2\privacyeps = \Omega(n^{-1})$, and the following holds: 
        \begin{equation*}
        \left\{\begin{array}{ll}
             \frac{n^{2/3}\accuracy^{1/3}}{\testingeps^{4/3}} + \frac{\sqrt{n}}{\testingeps^2} + \frac{\sqrt{n}}{\testingeps\sqrt{\privacyeps}} = O \left( \frac{\sqrt{n}\log n}{\sqrt{\privacyeps}} \right) &  \quaaad \text{and,}\\
             O \left( \frac{\sqrt{n}\log n}{\sqrt{\privacyeps}} \right) = O(n)& 
        \end{array}\right.
        \end{equation*}
        In this case, we run the augmented tester. The $\sqrt{n}\log n/\sqrt{\privacyeps}$ term dominates each term in~\cref{eq:augmentedclosenesscomplexity}. Therefore, we set $\ell = k = \Theta (\sqrt{n}\log n/\sqrt{\privacyeps})$, and
        \begin{equation*}
            s = \Theta \left( \frac{n}{\testingeps^2} \sqrt{\frac{2\accuracy}{k} + \frac{4}{n}}  + \frac{\sqrt{n}}{\testingeps\sqrt{\privacyeps}} \right) = O(k)
        \end{equation*}
        Similar to the previous case, this setting satisfies all three constraints. The sample complexity is $\Theta (\sqrt{n}\log n/\sqrt{\privacyeps})$. Since a $\sqrt{n}/(\testingeps\sqrt{\privacyeps})$ term appears in our lower bound, our sample complexity will be off by a factor of  $O(\testingeps\cdot\log n)$ compared to the lower bound.

        \item[Case 5:] $\testingeps = \Omega(n^{-1/4})$, $\testingeps^2\privacyeps = \Omega(n^{-1})$, and $\sqrt{n}\log n/\sqrt{\xi} = \Omega(n)$. In this case, we run the augmented tester. We set $\ell = k = \Theta(\log^2 n/\xi)$. Since $k = \Omega(n)$, we have $n''=O(k)$ and $\Delta(\overline{Z}) = \Theta((s+k)/k)$. Therefore, we require
        \begin{equation*}
            s = \Omega \left( \frac{k}{\testingeps^2}\sqrt{\frac{\alpha}{k} + \frac{1}{n}} + \frac{\sqrt{s+k}}{\testingeps\sqrt{\privacyeps}} \right) 
        \end{equation*}
        Notice that $\alpha/k = O(1/n)$, so solving for $s$ yields
        \begin{equation*}
            s = \Theta \left( \frac{k}{n\testingeps^2} + \frac{1}{\testingeps^2\privacyeps} + \frac{\sqrt{k}}{\testingeps\sqrt{\privacyeps}} \right) 
        \end{equation*}
        We can bound each of these terms using our assumptions. Since $\sqrt{n}\log n/\privacyeps = \Omega(n)$, we must have $\privacyeps = O( \log n / \sqrt{n})$. Using these facts, we can bound $s$:
        \begin{align*}
            \frac{k}{n\testingeps^2} &= O \left( \frac{\log^2 n}{n\testingeps^2\privacyeps} \right) = O(\log^2n) = O(k) \\   
            \frac{1}{\testingeps^2\privacyeps} &= O(n) = O(k) \\
            \frac{\sqrt{k}}{\testingeps\sqrt{\privacyeps}} &= \Theta \left( \frac{\log n}{\testingeps\privacyeps} \right) 
        \end{align*}
        Therefore, the overall sample complexity is $O(\log^2 n/\privacyeps + \log n / (\testingeps\privacyeps))$. Once again, this is within a factor of $O(\eps \cdot \log^2 n + \log n)$ of our lower bound. 
    \end{description}
\end{proof}

\subsection{Private Testing of $\ell_2$ Norm}
\label{sec:l2estimation}
To determine whether the two-step flattening worked, we must (privately) test the $\ell_2$ norm of the flattened distribution $p''$. The standard $\ell_2$ tester \citep{GR11, aliakbarpourIRS24} works by computing the number of collisions between elements in its sample set. This algorithm has the following guarantee:

 \begin{lemma}[\cite{GR11, aliakbarpourIRS24}] 
    Let $\errprob \in (0,1)$ and $p$ be a distribution over $[n]$. Let $E$ be a multiset of $\ell=O\left(\sqrt{n} \log(1/\errprob) \right)$ samples from $p$. For each $i \in [n]$, define $\ell_i$ as the number of instances of element $i$ in $E$. Let
    \begin{equation}
         L(E) = \frac{1}{{\ell \choose 2}} \sum_{i=1}^n {\ell_i \choose 2}
    \end{equation}
    Then with probability at least $1 - \errprob$,
    \begin{equation}
        \frac{\|p\|_2^2}{2} \leq L(E) \leq \frac{3\|p\|_2^2}{2}
    \end{equation}
\end{lemma}

The statistic above has high sensitivity; intuitively, if a large number of samples fall in the same bucket $(i,j,m)$, flipping a single instance of element $i$ could affect a large number of collisions. To avoid this, we will derandomize this tester by taking
\begin{equation}
    \label{eqn:derandomizedl2tester}
    \overline{L}(E) \coloneqq \mathbf{E}_{r} [L\mid E]
\end{equation}
where $r$ is the string of random bits used to choose the bucket when generating a sample from $p''$ given a sample from $p'$, and $E$ is a multiset (the \textit{estimation set}) of $\Poi{\ell}$ flattening samples from $p'$. Each of these samples is a pair $(i,j)$. Rather than sample the third coordinate to generate a sample from $p''$, we will consider the expected number of collisions over all such samplings, conditioned on $E$. In the result below, we give a closed-form expression for $\overline{L}$.

\begin{lemma}
\ifcolt
[Lemma~\ref{lemma:derandomizedl2tester}, restated]
\fi
    \label{lemma:derandomizedl2tester:repeat}
    Let $k_{i,j}$ and $\ell_{i,j}$ be the number of instances of element $(i,j)$ in the flattening and estimation sets $F$ and $E$ respectively. For each $i \in [n]$, let $b_i$ be the number of buckets created for element $i$ by the first flattening step. Then
    \begin{equation}
        \overline{L} = \frac{1}{{\ell \choose 2}} \sum_{i=1}^n \mathbf{E}_r \left[ \sum_{j=1}^{b_i} \sum_{k=1}^{b_{i,j}} {\ell_{i,j,k} \choose 2} \;\bigg\vert\; k_i, \ell_i \right] = \frac{1}{{\ell \choose 2}} \sum_{i=1}^n \sum_{j=1}^{b_i} \frac{{\ell_{i,j} \choose 2}}{k_{i,j} + 1}
    \end{equation}
\end{lemma}
\begin{proof}
    Fix $k_{i,j}$ and $\ell_{i,j}$. Recall that element $(i,j)$ is divided into $b_{i,j} = k_{i,j} + 1$ buckets. 
    Define $\ell_{i,j,m}$ as the number of instances of bucket $(i,j,m)$ in the estimation set. Since $b_{i,j}$ and $\ell_{i,j}$ are both fixed, $\ell_{i,j,m} \sim \textbf{Bin}(\ell_{i,j}, 1/b_{i,j})$, where the randomness comes only from $r$. We have the following:
    \begin{equation}
        \begin{aligned}
            \mathbf{E}_r[\ell_{i,j,m}] &= \frac{\ell_{i,j}}{b_{i,j}} \\
            \mathbf{E}_r[\ell_{i,j,m}^2] &= \textbf{Var}[\ell_{i,j,m}] + \mathbf{E}_r[\ell_{i,j,m}]^2 = \frac{\ell_{i,j}}{b_{i,j}}\left(1-\frac{1}{b_{i,j}}\right) + \frac{\ell_{i,j}^2}{b_{i,j}^2} = \frac{\ell_{i,j}}{b_{i,j}} + \frac{\ell_{i,j}^2 - \ell_{i,j}}{b_{i,j}^2}
        \end{aligned}
    \end{equation}
    We can now write the expectation as
    \begin{equation}
        \begin{aligned}
            \frac{1}{{\ell \choose 2}} \sum_{i=1}^n \mathbf{E}_r \left[ \sum_{j=1}^{b_i} \sum_{m=1}^{b_{i,j}} {\ell_{i,j,m} \choose 2} \;\bigg\vert\; k_i, \ell_i \right] 
            &= \frac{1}{{\ell \choose 2}} \sum_{i=1}^n \sum_{j=1}^{b_i} \sum_{m=1}^{b_{i,j}} \mathbf{E}_r \left[ \frac{\ell_{i,j,m}^2 - \ell_{i,j,m}}{2} \;\bigg\vert\; k_i, \ell_i \right] \\
            &= \frac{1}{{\ell \choose 2}} \sum_{i=1}^n \sum_{j=1}^{b_i} \sum_{m=1}^{b_{i,j}} \mleft(\frac{\ell_{i,j}}{2b_{i,j}} + \frac{\ell_{i,j}^2 - \ell_{i,j}}{2b_{i,j}^2} - \frac{\ell_{i,j}}{2b_{i,j}}\mright) \\
            &= \frac{1}{{\ell \choose 2}} \sum_{i=1}^n \sum_{j=1}^{b_i} b_{i,j} \cdot \frac{\ell_{i,j}^2 - \ell_{i,j}}{2b_{i,j}^2} \\
            &= \frac{1}{{\ell \choose 2}} \sum_{i=1}^n \sum_{j=1}^{b_i} \frac{\ell_{i,j}^2 - \ell_{i,j}}{2b_{i,j}} \\
            &= \frac{1}{{\ell \choose 2}} \sum_{i=1}^n \sum_{j=1}^{b_i} \frac{{\ell_{i,j} \choose 2}}{k_{i,j} + 1} 
        \end{aligned}
    \end{equation}
\end{proof}

In general, the sensitivity of $\overline{L}$ might be quite high. In the worst case, we might draw no instances of element $(i,j)$ in $F$ but $\Theta(\ell)$ instances in $E$, giving a sensitivity of $\Theta(1)$. However, note that the sensitivity is considerably lower if the flattening and estimation sets are similar (i.e., no element has a much higher frequency in $E$ than in $F$). The following lemma formalizes this intuition.  

\begin{lemma}
    \label{lemma:expectedsensitivity}
    Suppose that there exists $A \geq 0$ such that for all $i \in [n]$ and $j \in [b_i]$,
    \begin{equation}
        \label{eqn:boundedratios}
        \frac{\ell_{i,j}}{k_{i,j}+1} \leq A \cdot \frac{\ell}{k}
    \end{equation}
    Then the sensitivity of $\overline{L}$ is bounded by 
    \begin{equation}
        \label{eq:bound:sensitivity:under:assumption}
        \Delta(\overline{L}) \leq O \left( \frac{A^2}{k^2} + \frac{A}{k \ell} \right) 
    \end{equation}
\end{lemma}
\begin{proof}
    Assume that $X$ and $X'$ are fixed datasets which differ in only one sample (an instance of $(i,j)$ in $X$ is replaced by an instance of $(i',j')$ in $X'$). The extra instance can only change the contribution from $(i,j)$ and $(i',j')$. Let $k_{i,j}$ and $\ell_{i,j}$ represent the number of instances of $(i,j)$ in the the flattening and estimation sets of $X$ respectively. Let $k_{i,j}'$ and $\ell_{i,j}'$ represent the number of instances of $(i,j)$ in the flattening and estimation sets of $X'$. We will bound the difference for a single term; by symmetry, doubling the bound gives the overall sensitivity. 
    \begin{description}
        \item[Case 1:] The extra instance of $(i,j)$ falls in the flattening set $F$. In this case, $\ell_{i,j} = \ell_{i,j}'$ and $k_{i,j} = k_{i,j}'+1 \geq 1$. As $\ell_{i,j}/(k_{i,j}+1) \leq A \cdot \ell/k$ and $k_{i,j} \geq 1$,  $\ell_{i,j}/k_{i,j} \leq 2A \cdot \ell/k$. Therefore, 
        \begin{equation}
            \begin{aligned}
                \left| \frac{{\ell_{i,j} \choose 2}}{{\ell \choose 2} \cdot (k_{i,j} + 1)} - \frac{{\ell_{i,j}' \choose 2}}{{\ell \choose 2} \cdot (k_{i,j}' + 1)} \right|
                &= \left| \frac{{\ell_{i,j} \choose 2}}{{\ell \choose 2} \cdot (k_{i,j} + 1)} - \frac{{\ell_{i,j} \choose 2}}{{\ell \choose 2} \cdot (k_{i,j}' + 1)} \right| \\
                &= \frac{\ell_{i,j}^2 - \ell_{i,j}}{\ell^2 - \ell} \left| \frac{1}{k_{i,j}+1} - \frac{1}{k_{i,j}'+1} \right| \\
                &= \frac{\ell_{i,j}^2 - \ell_{i,j}}{\ell^2 - \ell} \left| \frac{k_{i,j}'+1 - (k_{i,j}+1)}{(k_{i,j}+1)(k_{i,j}'+1)} \right| \\
                &= \frac{\ell_{i,j}^2 - \ell_{i,j}}{\ell^2 - \ell} \cdot \frac{1}{k_{i,j}(k_{i,j}+1)} \\
                &\leq \frac{\ell_{i,j}^2}{\ell^2 - \ell} \cdot \frac{1}{k_{i,j}^2} \\
                &\leq \frac{1}{\ell^2 - \ell} \cdot \left( 2A \cdot \frac{\ell}{k} \right)^2 \\
                &\leq O \left( \frac{A^2}{k^2} \right)
            \end{aligned}
        \end{equation}
    
        \item[Case 2:] The extra instance of $i$ falls in the estimation set $E$, in which case $k_{i,j} = k_{i,j}'$ and $\ell_{i,j} = \ell_{i,j}' + 1$. Then we have
        \begin{equation}
            \begin{aligned}
                \left| \frac{{\ell_{i,j} \choose 2}}{{\ell \choose 2} \cdot (k_{i,j} + 1)} - \frac{{\ell_{i,j}' \choose 2}}{{\ell \choose 2} \cdot (k_{i,j}' + 1)} \right|
                &= \left| \frac{{\ell_{i,j} \choose 2}}{{\ell \choose 2} \cdot (k_{i,j} + 1)} - \frac{{\ell_{i,j}' \choose 2}}{{\ell \choose 2} \cdot (k_{i,j} + 1)} \right| \\
                &= \frac{\left|(\ell_{i,j}'+1)^2 - (\ell_{i,j}'+1) - \ell_{i,j}'^2 + \ell_{i,j}' \right|}{(\ell^2 - \ell)(k_{i,j}+1)} \\
                &= \frac{2\ell_{i,j}'}{(\ell^2 - \ell)(k_{i,j}+1)} \\
                &\leq \frac{2\ell_{i,j}}{k_{i,j}(\ell^2-\ell)} \\
                &\leq \frac{2A\ell}{k(\ell^2-\ell)} \\
                &\leq O \left( \frac{A}{\ell k} \right)
            \end{aligned}
        \end{equation}
    \end{description}
    Combining the two cases proves the lemma.
\end{proof}

For now, we assume~\cref{eqn:boundedratios} holds, and thus that the  sensitivity obeys the bound of~\cref{eq:bound:sensitivity:under:assumption} (in~\cref{sec:extendingdomain}, we show how to transform any dataset into one that satisfies this property for $A = \Theta(\log n)$). The next task is to show that under this assumption, we can estimate $\|p''\|_2^2$ to within a constant multiplicative factor.

\begin{lemma}
\ifcolt
[Lemma~\ref{lemma:privatel2tester}, restated]
\fi
    \label{lemma:privatel2tester:restated}
    Let $E$ be a set of $\Poi{\ell}$ estimation samples from $p$, and $\overline{L}$ be the derandomized $\ell_2^2$ norm estimator given in~\cref{eqn:derandomizedl2tester}. Let $A = 12\log(n/0.05)$. For a given value of $\privacyeps>0$, define
    \begin{equation}
        \tilde{L} = \overline{L} + \Lap{\frac{\Delta(L)}{\privacyeps}}
    \end{equation} 
    Suppose $\hat{k} =  \hat{k}_p +  \hat{k}_q\leq 100k$, that ~\cref{eq:bound:sensitivity:under:assumption} holds, and that
    \begin{equation}
        \label{eq:l2samplecomplexity}
        \begin{aligned}
            k \cdot \min(k/A^2,\ell/A) &\geq C_1\cdot \frac{k+n}{\privacyeps} \\
            \ell &\geq C_2 \sqrt{k+n}
        \end{aligned}
    \end{equation}
    for some sufficiently large constants $C_1,C_2$. Then
    \begin{equation*}
        \Pr {\left|\tilde{L} - \|p''\|_2^2 \right| > \frac{\|p''\|_2^2}{2} } \leq 0.06
    \end{equation*}
\end{lemma}
\begin{proof}
    First, we show that the magnitude of the Laplace noise is not too large. Since $\hat{k} \leq 100k$, the domain size of $p''$ is at most $n'' \leq 200k+3n$, and $\|p''\|_2 \geq 1/\sqrt{200k+3n}$. Using the pdf of the Laplace distribution and ~\cref{eq:l2samplecomplexity}, we have
    \begin{equation}
        \Pr{  \left|\tilde{L} - \overline{L} \right| > \frac{\|p''\|_2^2}{4} } = \exp \left( \frac{-\|p''\|_2^2 \privacyeps}{4\Delta(\overline{L})} \right) \leq \exp \left( \frac{-\privacyeps}{4(200k+3n)\Delta(\overline{L})} \right) \leq 1/100
    \end{equation}
    Next, we show that the derandomized statistic $\overline{L}$ concentrates around the true $\ell_2^2$ norm of $p''$. From \cite{GR11}, we have 
    \begin{equation}
        \E{\overline{L}} = \mathbf{E}_{X} \left[ \mathbf{E}_r[L] \right] =\mathbf{E}_{X,r} \left[ L \right] =  \|p''\|_2^2
    \end{equation}
    The variance of $\overline{L}$ can also be bounded using the result of \cite{GR11} and the law of total variance:
    \begin{equation}
        \Var{\overline{L}} = \Var{L} - \E{\Var{L \mid r}} \leq \Var{L} \leq 2 \left(\E{L}\right)^{3/2} = \frac{1}{{\ell \choose 2}^{1/2}} \cdot \left(\E{\overline{L}}\right)^{3/2} = \frac{\|p''\|_2^3}{{\ell \choose 2}^{1/2}}
    \end{equation}
    By Chebyshev's inequality, we obtain the following bound:
    \begin{equation}
        \label{eq:constraint:l}
        \Pr { \left| \overline{L} - \|p''\|_2^2 \right| > \frac{\|p''\|_2^2}{4} } \leq \frac{\Var{\overline{L}}}{\|p''\|_2^4/16} \leq \frac{32}{{\ell \choose 2}^{1/2} \cdot \|p''\|_2} \leq \frac{64\sqrt{200k+3n}}{\ell} \leq 0.05
    \end{equation}
    (The second to last inequality is true for all $\ell > 2$, and the last is true for $\ell \geq 1280 \sqrt{200k+3n}$.)
    
    Applying a union bound, the total error is bounded by 
    \begin{equation}
        \left| \tilde{L} - \|p''\|_2^2 \right| \leq \left| \tilde{L} - \overline{L} \right| + \left| \overline{L} - \|p''\|_2^2 \right| \leq \frac{\|p''\|_2^2}{4} + \frac{\|p''\|_2^2}{4} = \frac{\|p''\|_2^2}{2}
    \end{equation}
    with probability at least $0.94$, as desired. 
\end{proof}

\subsection{Extending the Domain of The Private Closeness Tester}
\label{sec:extendingdomain}

From Lemma \ref{lemma:expectedsensitivity}, it is clear that as long as Equation \ref{eqn:boundedratios} holds, the sensitivity is low. We will use the mapping technique given in \cite{ADKR19} to ensure that our dataset always has this property. Let $A \geq 2$. Define $\mathcal{X}$ as the set of all datasets (over domain $[n]$) and $\mathcal{X}^*$ as the subset of $\mathcal{X}$ which satisfies the property:
\begin{equation}
    \mathcal{X}^* = \left\{ X \in \mathcal{X} : \forall i \in [n], \frac{\ell_i}{k_i+1} \leq A \cdot \frac{\ell}{k} \right\}
\end{equation}
where, as before, $\ell_i$ and $k_i$ represent the number of occurrences of $i\in[n]$ in the testing and flattening parts of $X$, respectively. 
If $X \in \mathcal{X}^*$, we can add minimal noise to $\overline{L}$ and release a private statistic $\tilde{L}$ which is close to $L$ with high probability. The issue arises when $X \notin \mathcal{X}^*$. In this case, we use a mapping to transform $X$ into another dataset $Y \in \mathcal{X}$ such that the mapping from $X$ to $Y$ is differentially private. We do not worry about the tester's correctness in this case; the goal is merely to release a private statistic. 

\begin{lemma}
    \label{lemma:randomizedmapping}
    There exists a randomized mapping that takes $X, X' \in \mathcal{X}$ to $Y, Y' \in \mathcal{X}^*$ respectively with the following properties:
    \begin{itemize}
        \item If $X \in \mathcal{X}^*$, then $Y=X$
        \item If the Hamming distance between $X$ and $X'$ is 1, there exists a coupling $\mathcal{C}$ between the random outputs of the mapping $Y$ and $Y'$, where for any $(Y, Y') \sim \mathcal{C}$, the Hamming distance between $Y$ and $Y'$ is at most 4. 
    \end{itemize}
\end{lemma}
\begin{proof}
    The proof follows the idea of \cite[Lemma~C.5]{ADKR19}, with the required modifications to match our definition of $\mathcal{X}^\ast$. Given a flattening set $F$ and an estimation set $L$, the goal is to replace elements of $F$ until $\ell_i/(k_i+1) \leq A \cdot \ell/k$ for all $i$. For each element $i$, we need $r_i(X)$ extra copies in $F$, where 
    \begin{equation}
        r_i(X) = \max \left\{ \left\lceil \frac{k \cdot \ell_i(X)}{A \cdot \ell} \right\rceil - k_i(X) - 1, 0 \right\}
    \end{equation}
    Construct a multiset $R$ with $r_i(X)$ copies of $i$. To avoid violating the property, we must not remove more than $s_i(X)$ copies of each $i$, where 
    \begin{equation}
        s_i(X) = \max \left\{ k_i(X) + 1 -  \left\lceil \frac{k \cdot \ell_i(X)}{A \cdot \ell} \right\rceil, 0 \right\}
    \end{equation}
    To find these slots, mark $s_i$ instances of each $i$ as ``available" in $F$. We have at least $|R|$ available slots in total:
    \begin{equation}
        \begin{aligned}
            |R| &= \sum_{i=1}^n r_i(X) \leq \sum_{i=1}^n \frac{k \cdot \ell_i(X)}{A \cdot \ell} = \frac{k}{A} \leq (A - 1) \cdot \frac{k}{A} \leq k - \sum_{i=1}^n \frac{k_i(X)}{A} = k - \left( \sum_{i=1}^n \frac{k_i(X)}{A} \right) \\
            &\leq k - \left( \sum_{i=1}^n \left\lceil \frac{k_i(X)}{A} \right\rceil - 1 \right)
            \leq k - \sum_{i=1}^n k_i(X) - s_i(X) = \sum_{i=1}^n s_i(X)
        \end{aligned}
    \end{equation}
    We choose the first $|R|$ available slots in $F$ and insert elements of $R$ in those slots, randomly assigning elements to slots. The resulting dataset is in $\mathcal{X}^*$. If $|R| = 0$, then the original dataset was already in $\mathcal{X}^*$, and we change nothing. 

    Note that there are $|R|$ slots  and $|R|$ elements to be assigned to them, so there are $|R|!$ possible assignments of elements to slots. If we choose an assignment uniformly at random, each assignment has probability $1/|R|!$ of being chosen. The existence of the coupling then follows by a result shown as part of the proof of~\cite[Lemma~C.5]{ADKR19}:
    \begin{lemma}
        Let $\MM$ be the mapping above, and let $X, X'$ be two datasets which differ in exactly one sample. Then there exists a coupling $\CC$ between the outputs $\MM(X)$ and $\MM(X')$ such that for any $(\MM(X), \MM(X')) \sim \CC$, $\Ham{\MM(X)}{\MM(X')} \leq 4$. 
    \end{lemma}
\noindent This completes the proof.
\end{proof}

Given that our mapping satisfies Lemma~\ref{lemma:randomizedmapping}, there exists a choice of $A$ which decreases the error probability of the tester by at most $\errprob'$ while preserving privacy. 

\begin{lemma}[\cite{ADKR19}]
    \label{lem:composition}
    Let $\mathcal{A}$ be a $\privacyeps/4$-differentially private algorithm over $\mathcal{X}^*$ with parameter $A \geq 12 \log (n / \errprob')$ which tests the $\ell_2$ norm of $p$ and returns the correct answer with probability at least $1-\errprob$. Let $\mathcal{B}$ be a randomized mapping satisfying the conditions in Lemma \ref{lemma:randomizedmapping}. Then the algorithm which returns $\mathcal{A}(\mathcal{B}(X))$ is $\privacyeps$-differentially private and returns the correct answer with probability at least $1-\errprob-\errprob'$. 
\end{lemma}

\section{Lower Bound for Private Augmented Closeness Testing} \label{sec:LB_closeness}

We now state our lower bound for private augmented closeness testing. This result follows from the known lower bound for augmented closeness testing and a simple reduction from private identity testing.

\begin{theorem}
    \label{thm:closenesslowerbd}
    Let $\testingeps, \accuracy \in (0,1]$. Let $p$ and $q$ both be unknown distributions over $[n]$. Then any  $(\testingeps, \accuracy, \errprob=0.2)$-augmented closeness tester for $p$ and $q$ requires
    \begin{equation*}
        \Omega \left( \frac{n^{2/3}\accuracy^{1/3}}{\testingeps^{4/3}} + \frac{\sqrt{n}}{\testingeps^2} + \frac{n^{1/3}}{\testingeps^{4/3}\privacyeps^{2/3}} + \frac{\sqrt{n}}{\testingeps\sqrt{\privacyeps}} + \frac{1}{\testingeps\privacyeps} \right)
    \end{equation*}
    samples. 
\end{theorem}
\begin{proof}
The first two terms in the lower bound come directly from the lower bound for the non-private version of this problem:
\begin{lemma}[\cite{aliakbarpourIRS24}]
    \label{lem:augmentedclosenesslb}
    Let $\testingeps, \accuracy \in (0,1]$. Let $p$ and $q$ both be unknown distributions over $[n]$. Then, for every $\errprob\leq 11/24$, any  $(\testingeps, \accuracy, \errprob)$-augmented closeness tester for $p$ and $q$ requires
    \begin{equation*}
        \Omega \left( \frac{n^{2/3}\accuracy^{1/3}}{\testingeps^{4/3}} + \frac{\sqrt{n}}{\testingeps^2} \right)
    \end{equation*}
    samples. 
\end{lemma}
The proof of the remaining terms is via a standard reduction from 
the identity testing problem to the augmented closeness testing problem. A similar statement is provided in~\cite{aliakbarpourIRS24} in the non-private setting. Here we include the proof for the private version for the sake of completeness.

    Fix $\delta \leq 1/5$. Given a $(\privacyeps, \testingeps, \accuracy, \delta)$-private augmented closeness tester and an instance of identity testing between unknown distribution $p$ and known distribution $q$, we set $\hatp = q$ and run the closeness tester with samples from $p$ and $q$, predicted distribution $\hatp$, and suggested accuracy $\alpha = \testingeps$. If the closeness tester returns $\reject$ or $\invalid$, we output $\reject$; if the closeness tester returns $\accept$, we output $\accept$. 

    Next, we show that with high probability our output is correct. If $\totalvardist{p}{q} > \testingeps$, the closeness tester will not return $\accept$ with probability greater than $1/5$. If $p = q$, since $q = \hatp$, the closeness tester will not return $\invalid$ or $\reject$ with probability greater than $2\delta \leq 2/5$ by a union bound. Finally, the closeness tester is $\privacyeps$-differentially private, so any function of its output is similarly private. Therefore, the closeness tester can be used to construct a $\privacyeps$-private $(\testingeps, 2\delta)$-identity tester, and the lower bound of \cite{AcharyaSZ18} stated in Lemma~\ref{lemma:privateidtesting} gives a lower bound for private augmented closeness testing. 
\end{proof}

\paragraph{Acknowledgments}This work was initiated while MA was serving as a research fellow and CC and RR were visiting the Simons Institute for the Theory of Computing as part of the Sublinear Algorithms program.

\printbibliography

\appendix

\end{document}